\documentclass[lettersize,journal]{IEEEtran}
\usepackage{amsmath,amsfonts}
\usepackage[ruled,vlined]{algorithm2e}
\usepackage{array}
\usepackage[caption=false,font=normalsize,labelfont=sf,textfont=sf]{subfig}
\usepackage{textcomp}
\usepackage{stfloats}
\usepackage{url}
\usepackage{verbatim}
\usepackage{graphicx}
\usepackage{cite}
\usepackage{multirow}
\usepackage{amsmath,amssymb}
\usepackage{booktabs}       
\usepackage{array}          
\usepackage{xspace}

\usepackage{tikz}
\usetikzlibrary{positioning, fit, backgrounds, calc, arrows.meta, shapes.geometric}

\newcommand{\name}{\textsc{Trace}\xspace}
\newcommand{\prism}{\textsc{Prism}\xspace}
\newcommand{\astra}{\textsc{Astra}\xspace}
\newcommand{\voltage}{\textsc{Voltage}\xspace}
\newcommand{\Kvq}{\ensuremath{\mathbf{K}_{\mathrm{vq}}}\xspace}
\newcommand{\Vvq}{\ensuremath{\mathbf{V}_{\mathrm{vq}}}\xspace}

\begin{document}

\title{TRACE: Transformer Residual Activation Codebook Encoding for Communication-efficient Distributed Edge Inference}

\author{Muhammad Azlan Qazi, Alexandros Iosifidis, and Qi Zhang
\thanks{Muhammad Azlan Qazi and Qi Zhang are with DIGIT and the Department of Electrical and Computer Engineering, Aarhus University, Denmark (e-mail: maq@ece.au.dk; qz@ece.au.dk)}
\thanks{Alexandros Iosifidis is with the Data Science Research center, Tampere University, Finland (e-mail: alexandros.iosifidis@tuni.fi)}
}


\maketitle

\begin{abstract}
Distributed Transformer inference across co-located edge devices is fundamentally communication-bound, as exchanging activations between sequence-parallel partitions dominates inference latency under commodity edge bandwidths. Existing compression schemes reduce this cost but retain a structural bottleneck: the received data must still undergo key/value projection before attention can proceed, serializing computation behind communication. This paper presents Transformer Residual Activation Codebook Encoding (\name{}), a framework that addresses these limitations. \name{} places a Vector Quantizer (VQ) at the post-attention residual of every Transformer layer and precomputes codebook-indexed key/value representations. During inference, exchanged activations are encoded into compact codebook indices, allowing receivers to directly recover keys and values through table lookup instead of executing the downstream computations required to derive them. By eliminating the received data downstream computation, \name{} enables communication--computation overlap and reduces a key latency bottleneck in distributed Transformer inference.
A hierarchical clustering further structures the learned codebook into a multi-resolution hierarchy, enabling runtime adaptation of the communication bitrate without retraining. The framework supports two operating variants: one prioritizing accuracy and the other balancing accuracy and efficiency through buffer-compatible quantization. Experiments on ViT-Base (ImageNet-1K and CIFAR-100) and ViViT-B (UCF-101) show that \name{} achieves up to $1.89\%$ higher accuracy and up to $1.27\times$ end-to-end inference speedup  compared with the state-of-the-art VQ-based distributed inference framework. Compared with uncompressed sequence-parallel inference, \name{} reduces communication overhead by up to $22{,}624\times$.
\end{abstract}


\section{Introduction}
\label{sec:intro}

\IEEEPARstart{T}{ransformer} architectures have emerged as the dominant computational paradigm across vision, language, and multimodal artificial intelligence. Trained at scale, they now define state-of-the-art performance across perception, understanding, and generation tasks~\cite{dosovitskiy2020image,arnab2021vivit,yang2025qwen25,touvron2023llama,touvron2023llama2,guo2025deepseek,google2025veo3launch}. Their accuracy, however, comes at a steep computational and memory price: a single forward pass through a state-of-the-art Transformer model involves billions of floating-point operations and hundreds of megabytes of intermediate activations, making single-device deployment on resource-constrained edge nodes impractical.

Centralized cloud serving is increasingly unsuitable for IoT workloads. End-user latency targets, privacy guarantees, energy budgets, and unreliable wide-area connectivity all motivate to bring inference at the network edge~\cite{10818760,li2025tpi,ye2025jupiter}. When a single edge device cannot host the model alone, the natural answer is \emph{collaborative} or \emph{distributed} inference, in which a model is partitioned across multiple co-located edge nodes whose aggregate compute and memory cover the model's demands~\cite{10631032,jiang2023hexgen,peccia2024embedded}. Three parallelism families have been developed for this purpose, namely tensor parallelism in Megatron-LM~\cite{shoeybi2019megatron}, pipeline parallelism in GPipe~\cite{huang2019gpipe}, and sequence parallelism~\cite{li-etal-2023-sequence}. All three were designed for high-bandwidth datacenter interconnects such as NVLink, where all-reduce and ring activation transfers consume negligible time.

The picture changes dramatically at edge bandwidths. In a sequence-parallel deployment with hidden dimension $D$, sequence length $N$, and $P$ participating devices, every Transformer layer must exchange $(P{-}1)ND/P$ floating-point values per device per layer before the next layer can proceed~\cite{10631032}. At commodity edge bandwidths of tens to a few hundreds Mbps, this volume implies per-layer communication latencies of tens to hundreds of milliseconds, which is several orders of magnitude above local computation time~\cite{anthony2024demystifying,pope2023efficiently,10546617}. Empirical measurements confirm that under bandwidth constraints below $100$~Mbps, inter-device communication accounts for roughly $60$--$93\%$ of total inference latency in distributed BERT execution ($N{=}1024$, $P{=}4$), largely negating the speed-up that parallelism is supposed to provide~\cite{liu2025communication}.

\begin{figure*}[t]
  \centering
  \includegraphics[width=\textwidth]{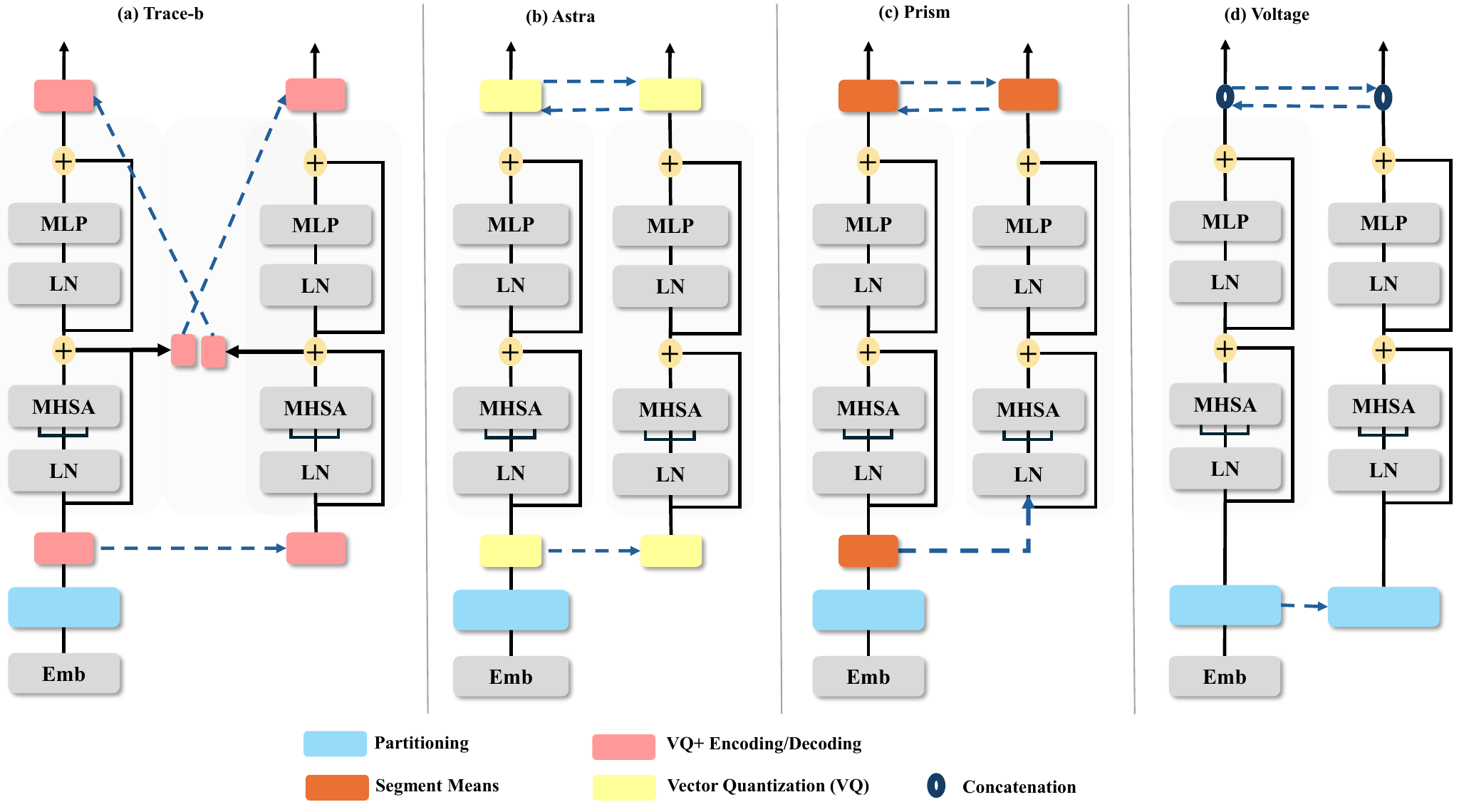}
  \caption{%
    Comparison of distributed Transformer inference frameworks across two sequence-parallel devices. 
	\textbf{(a)~\name-b (ours)}: transmits only a codebook index; the receiver fetches precomputed $\mathbf{K}$/$\mathbf{V}$ entries from a buffer and concatenates them with local keys and values, eliminating receiver-side projection and enabling full compute/communication overlap.
	\textbf{(b)~\astra}~\cite{liu2025communication}: quantizes pre-attention embeddings via VQ; receiver-side projection remains a hard dependency. 
	\textbf{(c)~\prism}~\cite{qazi2025prism}: transmits Segment Means (token-group averages); receiver still applies $\mathbf{K}$/$\mathbf{V}$ projections before attention. 
	\textbf{(d)~\voltage}~\cite{10631032}: transmits the full hidden-state partition; no compression. 
  }
  \label{fig:methods}
\end{figure*}

To address this bottleneck, several lines of work directly compress the communicated payload. Theoretical results in distributed inference~\cite{9459473} establish that combined sparsification and quantization preserve convergence at extremely low bit budgets, providing the foundation for compressed inter-device communication. \voltage~\cite{10631032}, illustrated in Fig.~\ref{fig:methods}(d), is the baseline sequence-parallel design: each device transmits its full-precision hidden-state partition to peers at the partitioning boundary, with communication volume $(P{-}1)ND/P$ per layer and no compression of any kind. \prism~\cite{qazi2025prism}, shown in Fig.~\ref{fig:methods}(c), takes a significant step by replacing this full transmission with training-free \emph{Segment Means} (column-wise averages of token sub-groups) substantially reducing communication at the cost of approximation accuracy. \astra~\cite{liu2025communication}, shown in Fig.~\ref{fig:methods}(b), applies vector quantization (VQ) \emph{inside} multi-head self-attention, quantizing pre-attention token embeddings exchanged between devices and recovering them on the receiver for direct use in the attention kernel.

\astra  and \prism share a common structural limitation visible in Figs.~\ref{fig:methods}(b)--(c), i.e., both still require the receiver to complete $\mathbf{K}$/$\mathbf{V}$ projection step before attention can proceed, preventing full overlap of communication with downstream computation (Section~\ref{sec:background}). \prism's per-token payload also remains a $D$-dimensional float vector, so communication volume scales with model dimension regardless of compression ratio. \astra's bitrate is fixed at finetuning time with no mechanism for adaptive resolution.

This paper proposes \textbf{T}ransformer \textbf{R}esidual \textbf{A}ctivation \textbf{C}odebook \textbf{E}ncoding (\name), a communication-efficient framework for distributed Transformer inference that closes both gaps simultaneously. The core idea is to place VQ at the post-attention layer boundary (on the hidden state $\mathbf{Z}$ produced by the multi-head self-attention sublayer and its first residual addition) and to amortize the next layer's $\mathbf{K}$ and $\mathbf{V}$ projections into precomputed lookup buffers indexed by codebook entry, so that the receiver recovers cross-device quantized $\mathbf{K}$/$\mathbf{V}$ via direct table lookup with zero matrix-multiply computation. A hierarchical $k$-medoids clustering~\cite{kaufman2009finding} of the trained codebook then exposes a multi-resolution sub-codebook hierarchy, allowing the deployed model to select bitrate at inference time without retraining. It is imperative to mention that, this paper targets the encoder-only architectures used for perception and understanding tasks (e.g., ViT, ViViT). Extending the mechanism to autoregressive decoder-only LLMs is discussed as future work in Section~\ref{sec:discussion}.

Across all settings, \name closely matches baseline accuracy at significant communication overhead reduction compared with uncompressed sequence-parallel inference methods, and is also superior to the best prior compression methods while enabling seamless compute-communication overlap that those methods cannot provide.

The main contributions of this work are:




\begin{itemize}

\item \textbf{Two accuracy--latency variants.}
\name offers two variants that differ in the placement of vector quantization relative to the post-attention residual (Section~\ref{sec:eval:ablation}). \name-nb quantizes \emph{before} the residual and retains runtime-dependent residual information for higher accuracy, whereas \name-b quantizes \emph{after} the residual, enabling deterministic mapping to precomputed key/value entries and compute--communication overlap.

\item \textbf{Post-attention VQ with precomputed K/V buffers.}
The hidden state $\mathbf{Z}$ at the post-attention boundary is quantized using a learned codebook, while the downstream layer normalization, MLP, and $\mathbf{K}/\mathbf{V}$ representations of each codeword are precomputed during post-training calibration. At inference, only codebook indices are transmitted and cross-device $\mathbf{K}/\mathbf{V}$ representations are recovered by direct table lookup, eliminating receiver-side matrix multiplication for reconstruction.

\item \textbf{Communication--computation overlap.}
In \name-b, devices transmit codebook indices while concurrently executing local MLP and $\mathbf{K}/\mathbf{V}$ computations on their available partition data. This removes the receiver-side projection dependency on incoming activations, avoiding the communication--computation serialization that limits prior sequence-parallel methods.

\item \textbf{Multi-resolution bitrate adaptation.}
\name constructs a hierarchical multi-resolution codebook using $k$-medoids clustering~\cite{kaufman2009finding} on the trained codebook. Each hierarchy level forms a valid sub-codebook, enabling runtime bitrate selection without retraining and adaptation to changing bandwidth and resource constraints.

\item \textbf{Modal evaluation.}
\name is evaluated on image classification (ViT-Base/16 on ImageNet-1K and CIFAR-100) and video classification (ViViT-B/16 on UCF-101). \name-nb improves top-1 accuracy by up to $1.89\%$ over the strongest prior VQ-based method, while \name-b achieves up to $1.27\times$ faster inference with up to $1.45\%$ higher accuracy; both variants reduce transmitted activation payloads by up to $22{,}624\times$ for ViT-Base/16 and $24{,}500\times$ for ViViT-B/16 compared with uncompressed sequence-parallel inference.

\end{itemize}

The remainder of this paper is organized as follows. Section~\ref{sec:background} reviews the relevant background and related work. Section~\ref{sec:system-overview} provides a high-level overview of \name and its operational lifecycle. Section~\ref{sec:method-detail} presents the technical details of codebook training, K/V buffer calibration, distributed inference, and multi-resolution operation. Section~\ref{sec:eval} describes the experimental setup and results. Section~\ref{sec:discussion} discusses the implications, limitations, and future directions, and Section~\ref{sec:conclusion} concludes the paper.

\noindent\textit{Notation.} Bold uppercase ($\mathbf{X}$) and lowercase ($\mathbf{x}$) denote matrices and vectors, respectively. Italic lowercase ($k$, $n$) denotes scalars or indices; italic uppercase ($N$, $D$, $P$) and Greek letters ($\gamma$, $\sigma$) denote constants or hyperparameters. Calligraphic letters ($\mathcal{I}$) denote sets, and $\mathbb{R}$ the reals. \textit{Device}, peers and \textit{edge device} are used interchangeably.

\begin{figure*}[t]
    \centering
    \includegraphics[width=0.8\textwidth]{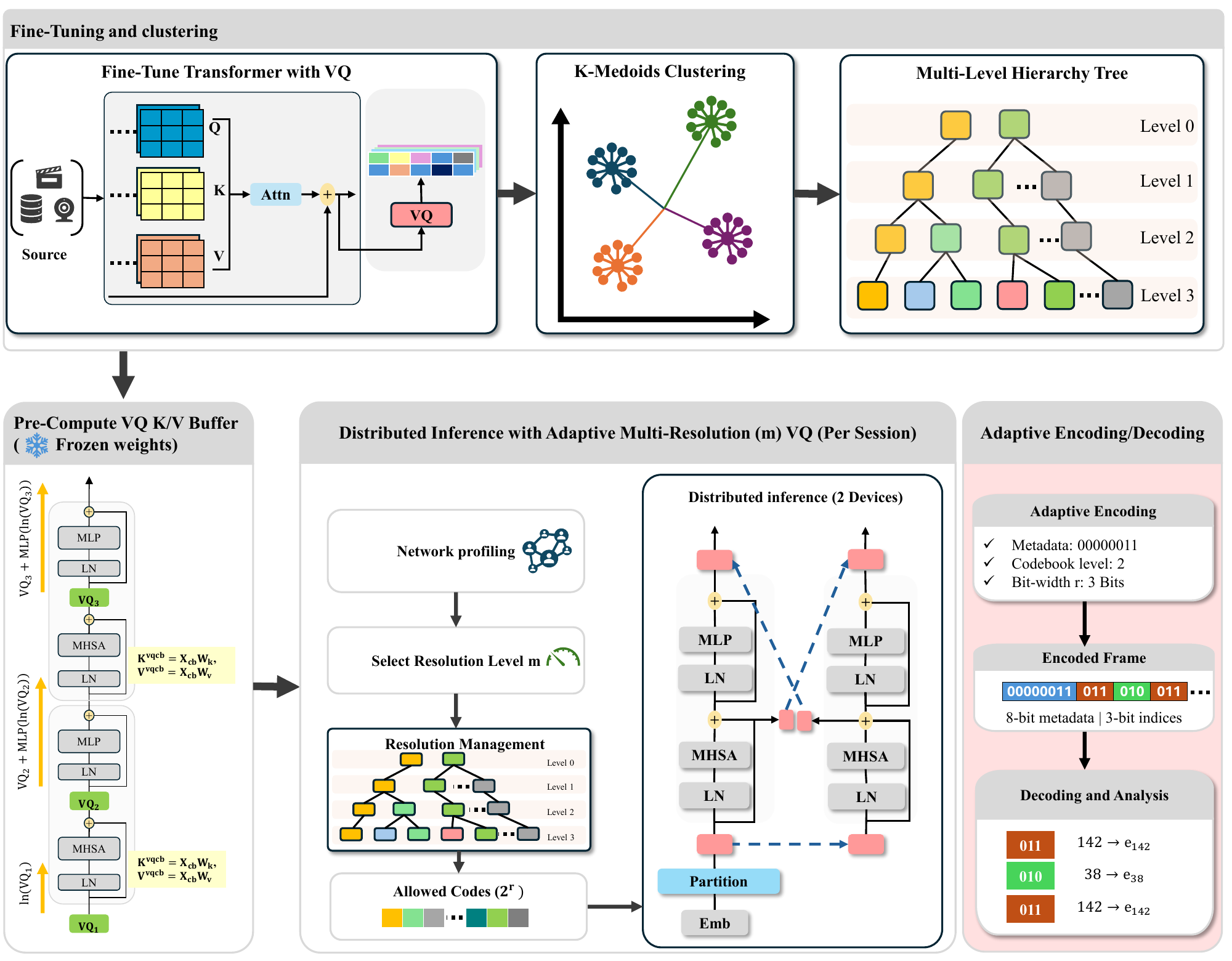}
    \caption{
        \textbf{\name system overview.} (\textit{Top}) Offline pipeline: finetune Transformer with VQ  modules, build a multi-level hierarchy tree via hierarchical  k-medoids clustering.
         (\textit{Bottom-left}) Pre-compute per-layer  $\mathbf{K}^\mathsf{vqcb}$ and $\mathbf{V}^\mathsf{vqcb}$  buffers with frozen weights.
         (\textit{Bottom-center}) Per-session adaptive inference:  select resolution level $m$ from network conditions, exchange  $r$-bit indices between devices overlapped with local FFN  computation.
         (\textit{Bottom-right}) Adaptive encoding with 8-bit metadata  header; receiver decodes indices and gathers precomputed K/V  buffers --- no MatMul on cross-device tokens.
    }
    \label{fig:trace_architecture}
\end{figure*}

\section{Background \& Related Work}
\label{sec:background}

\subsection{Transformer Architecture}
\label{sec:bg:Transformer}

The Transformer~\cite{46201} replaces recurrent and convolutional sequence models with a fully attention-based design. ViT~\cite{dosovitskiy2020image} extended the architecture to vision by treating fixed-size image patches as a token sequence. DeiT~\cite{touvron2021training} subsequently improved data efficiency through Transformer-specific knowledge distillation. ViViT~\cite{arnab2021vivit} generalized the design to video by extracting spatio-temporal tube tokens and applying spatial--temporal attention. BERT~\cite{devlin2019bert} established the encoder-only bidirectional Transformer paradigm for natural language understanding, while the autoregressive decoder family, including modern LLMs such as Qwen2.5~\cite{yang2025qwen25}, DeepSeek-R1~\cite{guo2025deepseek}, and LLaMA~\cite{touvron2023llama,touvron2023llama2}, has redefined the state of the art on language tasks. \name is designed to apply distributed inference across encoder (ViT, ViViT) variants.

The core building block in all of the Transformer based models is the multi-head self-attention (MHSA) sublayer. For each of $H$ heads, the input sequence $\mathbf{X}_p^{(\ell)} \in \mathbb{R}^{N_p \times D}$ of device $p$ at layer $\ell$ is first projected into queries, keys, and values via three learned weight matrices $\mathbf{W}_Q$, $\mathbf{W}_K$, $\mathbf{W}_V \in \mathbb{R}^{D \times d}$:
\begin{equation}
    \mathbf{Q}_p^{(\ell)} = \mathbf{X}_p^{(\ell)}\mathbf{W}_Q, \quad
    \mathbf{K}_p^{(\ell)} = \mathbf{X}_p^{(\ell)}\mathbf{W}_K, \quad
    \mathbf{V}_p^{(\ell)} = \mathbf{X}_p^{(\ell)}\mathbf{W}_V,
    \label{eq:qkv}
\end{equation}
where $\mathbf{Q}_p^{(\ell)}, \mathbf{K}_p^{(\ell)}, \mathbf{V}_p^{(\ell)} \in \mathbb{R}^{N_p \times d}$ and $d = D/H$ is the per-head dimension. Scaled dot-product attention is then applied:
\begin{equation}
    \operatorname{Attention}( \mathbf{Q}_p^{(\ell)}, \mathbf{K}_p^{(\ell)}, \mathbf{V}_p^{(\ell)})
    = \operatorname{softmax}\!\left(
        \frac{\mathbf{Q}_p^{(\ell)}{\mathbf{K}_p^{(\ell)}}^{\top}}{\sqrt{d}}
      \right) \mathbf{V}_p^{(\ell)}.
    \label{eq:attn}
\end{equation}
MHSA is followed by a position-wise feed-forward sublayer (MLP), and both sublayers are wrapped with residual connections and layer normalization, giving each Transformer layer the structure
\begin{align}
    \mathbf{Z}_p^{(\ell)} &= \mathbf{X}_p^{(\ell)}
        + \operatorname{MHSA}\!\left(
          \operatorname{LN}_1\!\left(\mathbf{X}_p^{(\ell)}\right)\right),
        \label{eq:zdef}\\
    \mathbf{Y}_p^{(\ell)} &= \mathbf{Z}_p^{(\ell)}
        + \operatorname{MLP}\!\left(
          \operatorname{LN}_2\!\left(\mathbf{Z}_p^{(\ell)}\right)\right).
        \label{eq:ydef}
\end{align}
The intermediate tensor $\mathbf{Z}_p^{(\ell)}$ in~\eqref{eq:zdef} is the activation that \name quantizes, encodes, and transmits, making it the central object of the proposed framework.

\subsection{Distributed Transformer Inference at the Edge}
\label{sec:bg:parallelism}
\label{sec:bg:edge}

Distributing Transformer inference across multiple devices has been studied primarily in datacenter settings. \emph{Tensor parallelism}, introduced in Megatron-LM~\cite{shoeybi2019megatron}, partitions individual weight matrices and inserts an all-reduce collective after each MHSA and MLP sublayer, exchanging hidden-state shards across devices at every sublayer boundary. \emph{Pipeline parallelism}, exemplified by GPipe~\cite{huang2019gpipe}, assigns contiguous groups of layers to different devices and processes micro-batches in a pipeline, passing the full hidden state at each stage transition. \emph{Sequence parallelism}~\cite{li-etal-2023-sequence,fang2024usp} distributes the token sequence ($\mathbf{X}^{(\ell)} \in \mathbb{R}^{N \times D}$) across devices. Each device computes Transformer block for available partitioned sequence $\mathbf{X}_p^{(\ell)} \in \mathbb{R}^{N_p \times D}$, and communicates respected output $\mathbf{Y}_p^{(\ell)}$ to other devices. This cycle is repeated for the rest of the Transformer blocks. Pope~\textit{et~al.}~\cite{pope2023efficiently} provide a systematic analysis confirming that, in all three families, inter-device communication is the principal bottleneck at inference stage.

This bottleneck becomes even more acute at the network edge, where a growing body of work targets Transformer inference under tighter resource constraints. DeTransformer~\cite{10546617} restructures each Transformer layer into decoupled sub-blocks and adaptively places them across edge devices, motivating bandwidth-aware design. EdgeShard~\cite{10818760} addresses LLM partitioning across heterogeneous edge--cloud nodes through a joint device-selection and shard-placement formulation. TPI-LLM~\cite{li2025tpi} demonstrates serving 70B-parameter models on low-resource edge devices via star-based scheduling and memory windowing. STI~\cite{guo2023sti} elastically pipelines parameter loading and computation on a single mobile device, while Jupiter~\cite{ye2025jupiter} extends collaborative LLM inference to recent-generation edge hardware. RCIF~\cite{10446903} specifically targets robustness under the highly lossy network conditions characteristic of many IoT deployments, a setting in which \name's ultra-low-bitrate index transmission provides additional resilience against packet loss relative to floating-point transmissions. For broader context on distributed neural-network inference across embedded and IoT systems, see the recent systematic review in~\cite{peccia2024embedded}.

The works most directly related to \name are \voltage, \prism, and \astra, all of which target sequence-parallel distributed Transformer inference but differ in what they compress and where. \voltage~\cite{10631032} is an edge-targeted distributed inference system that exploits position-wise sequence partitioning and adapts the partition strategy to measured bandwidth and device speed. It serves as the uncompressed baseline, transmitting the full $D$-dimensional hidden-state partition between devices at every layer boundary with communication volume $\mathcal{O}(N \times D)$. \prism~\cite{qazi2025prism} replaces full transmission with training-free Segment Means (column-wise averages of token sub-groups) achieving substantial communication reduction across ViT, BERT, and GPT-2. The receiving device extends its local key and value matrices with the received Segment Means features and applies a scaling-aware softmax correction. Two limitations remain: the transmitted payload is still a $D$-dimensional float vector per segment, and the receiver must apply $\mathbf{W}_k$/$\mathbf{W}_v$ projection matrix multiplications to those vectors before attention can proceed. \astra~\cite{liu2025communication} goes further by using learned VQ codewords in place of Segment Means. Each device quantizes its pre-attention hidden features against a shared codebook, transmits the assigned codeword indices to its peers, and likewise receives indices for the partitions held by other devices. The received indices are de-referenced through the codebook to reconstruct the corresponding quantized embeddings, from which non-local keys and values $\Kvq$/$\Vvq$ are then computed via the $\mathbf{W}_k$/$\mathbf{W}_v$ projection matrices. Each device finally concatenates these reconstructed $\Kvq$/$\Vvq$ with the keys and values of its own local partition and performs attention over the combined sequence. However, because VQ is placed before attention, the receiver must still apply the $\mathbf{W}_k$/$\mathbf{W}_v$ projection matrix multiplications to every quantized embedding before attention can proceed. This is a non-trivial compute dependency that limits compute--communication overlap and creates a data synchronization bottleneck. \astra's codebook size and bits per vector are also fixed at training time, requiring retraining to adapt to varying link and codebook setups.

\subsection{Vector Quantization}
\label{sec:bg:vq}

Vector quantization (VQ)~\cite{10.5555/128857} represents a continuous
vector $\mathbf{z} \in \mathbb{R}^D$ by the index of its nearest entry
in a learned codebook
$\mathcal{C} = \{\mathbf{c}_1, \ldots, \mathbf{c}_K\} \subset \mathbb{R}^D$:
\begin{equation}
    k = \operatorname*{arg\,min}_j \|\mathbf{z} - \mathbf{c}_j\|_2,
    \qquad
    q(\mathbf{z}) = \mathbf{c}_{k}.
    \label{eq:vq}
\end{equation}
Each vector is transmitted as a single codeword index of $\lceil \log_2 K \rceil$ bits, the minimum number of bits required to address one of $K$ codewords. This is determined solely by the codebook size $K$ and independent of the feature dimension $D$, yielding compression ratios that grow favorably with $D$. However, nearest-centroid search in high-dimensional spaces is susceptible to codebook collapse as $D$ grows~\cite{van2017neural}. In \name\ this is mitigated by cosine-similarity assignment, $k$-means initialization on real activations, and exponential moving average EMA-based dead-code recycling, as described in Section~\ref{sec:method:training}.

VQ-VAE~\cite{van2017neural} established the standard framework for end-to-end neural VQ training. A straight-through estimator (STE) enables gradient propagation through the non-differentiable assignment, a commitment loss encourages encoder outputs to remain close to their assigned codebook entries (codewords), and an EMA update rule refines the codebook vectors directly from encoder statistics improving training stability compared to purely gradient-based codebook updates.

Multi-stage designs such as Group-Residual VQ in HiFi-Codec~\cite{yang2023hifi} stack codebooks to iteratively reduce quantization error, achieving variable-bitrate compression by varying the number of residual stages. In contrast, \name derives variable bitrate from a hierarchical sub-codebook configuration of a \emph{single} codebook. This preserves a single-index representation per vector, enabling efficient precomputation of the corresponding quantized representations from the trained codebook for fast communication-time reconstruction.


\section{System Overview}
\label{sec:system-overview}

\name targets sequence-parallel Transformer inference across $P$ co-located edge devices. As illustrated in Fig.~\ref{fig:trace_architecture}, the input sequence is partitioned position-wise across devices, while each device stores a copy of the model. Device $p$ processes a local partition $\mathbf{X}_p^{(\ell)} \in \mathbb{R}^{N_p \times D}$, where $N_p=N/P$. Global self-attention requires each device to access the representations associated with peer partitions, making this inter-device exchange the main communication bottleneck addressed by \name.

The central idea of \name is to quantize the post-attention residual of each Transformer layer and communicate only the resulting codebook indices. For every codeword, \name stores the precomputed corresponding K/V representations. Upon receiving the peer device's encoded representation, the device decodes it and maps directly to quantized K/V vectors through buffer retrieval, without applying the intermediate neural transformations (MLP) and K/V projections to the received data. This allows the communication to proceed concurrently with local computation. This mechanism defines the default variant \name-b. We also proposed a second variant \name-nb which forgoes buffer retrieval in exchange for higher accuracy. Both are defined in Section~\ref{sec:system-dataflow}. Unless stated otherwise, \name refers to \name-b.

\subsection{Lifecycle}
\label{sec:system-lifecycle}

\name consists of three phases. First, during offline finetuning, a layer-specific vector quantizer is inserted at each post-attention residual. An input-stage quantizer $\mathrm{VQ}_0$ is also used so that the first inter-device exchange is represented by codebook indices. The model and codebooks are jointly finetuned for the target task.

Second, after finetuning, the model weights and codebooks are frozen. Each trained codebook is organized into a nested hierarchy of sub-codebooks using hierarchical $k$-medoids clustering. This provides multiple resolution levels from a single trained model, allowing the communication data size to be selected according to the available bandwidth, latency budget, and desired accuracy. A one-time calibration procedure then computes, for every codeword, the K/V representations required by the subsequent attention layer and stores them in the corresponding $\mathbf{K}^{\mathsf{vqcb}}$ and $\mathbf{V}^{\mathsf{vqcb}}$ lookup buffers. Since each sub-codebook contains entries from the original trained codebook, all resolution levels reuse the same buffers without recalibration.

Third, during online inference, an active resolution level is selected and each device quantizes its local post-attention residual using the corresponding sub-codebook. The device transmits only the resulting indices, which its peers use to retrieve the corresponding rows from the precomputed K/V buffers. These non-local K/V representations are combined with the locally computed K/V representations for global attention. The finetuning procedure, hierarchy construction, buffer calibration, and distributed inference operations are detailed in Section~\ref{sec:method-detail}.

\subsection{Online Data Flow and Runtime Resolution}
\label{sec:system-dataflow}


The bottom-center panel of Fig.~\ref{fig:trace_architecture}  illustrates the per-layer data flow described in this subsection. Consider a single layer transition from $\ell$ to $\ell{+}1$ on  device $p$, with peers indexed $p' \neq p$. Five operations occur per layer (operations~2 and~3 execute in parallel):

\begin{enumerate}
    \item \textbf{Local layer execution.}
    Device $p$ computes $\mathbf{Z}_p^{(\ell)}$ over its own partition via Eq.~\eqref{eq:zdef}.
    
    \item \textbf{Quantization and index broadcast.}
    Each token vector $\mathbf{z}_{p,n}^{(\ell)}$ is assigned to its nearest entry in the active resolution-level $m$ sub-codebook by cosine similarity (Eq.~\ref{eq:vq-insert}), and the resulting local sub-codebook indices $\{i_{p,n}^{(\ell)}\}_{n=1}^{N_p}$ are packed into $r$-bit fields with an 8-bit metadata header and broadcast to all peers.

    \item \textbf{Local MLP and projections.}
    In parallel with the broadcast, device $p$ continues local computation: $\mathbf{Y}_p^{(\ell)} = \mathbf{Z}_p^{(\ell)} + \mathrm{MLP}(\mathrm{LN}_2(\mathbf{Z}_p^{(\ell)}))$, followed by the layer-$\ell{+}1$ K/V projections on the local partition to obtain $\mathbf{K}_p^{(\ell+1)}$, $\mathbf{V}_p^{(\ell+1)}$.

 \item \textbf{Cross-device K/V lookup.}
    Once the local indices $\{i_{p',n}^{(\ell)}\}$ from peer $p'$ arrive, device $p$ maps each to its global codebook position $g_{p',n} = \boldsymbol{\pi}[i_{p',n}^{(\ell)}]$ (Section~\ref{sec:method:multires}) and retrieves the corresponding rows of $\mathbf{K}^{\mathsf{vqcb}}_{\ell+1}$ and $\mathbf{V}^{\mathsf{vqcb}}_{\ell+1}$ by direct array indexing.
      \item \textbf{Split-attention.}
    Device $p$ concatenates its local $\mathbf{K}_p^{(\ell+1)}$/$\mathbf{V}_p^{(\ell+1)}$ with the buffer-recovered cross-device sets $\{\mathbf{K}^{\mathrm{vq},(\ell+1)}_{p'}\}_{p'\neq p}$/ $\{\mathbf{V}^{\mathrm{vq},(\ell+1)}_{p'}\}_{p'\neq p}$ (Eq.~\ref{eq:concat-k}) to form global key/value matrices, and computes attention with its local queries $\mathbf{Q}_p^{(\ell+1)}$ over this concatenation.
\end{enumerate}

\begin{figure*}[t]
    \centering
    \includegraphics[width=\textwidth]{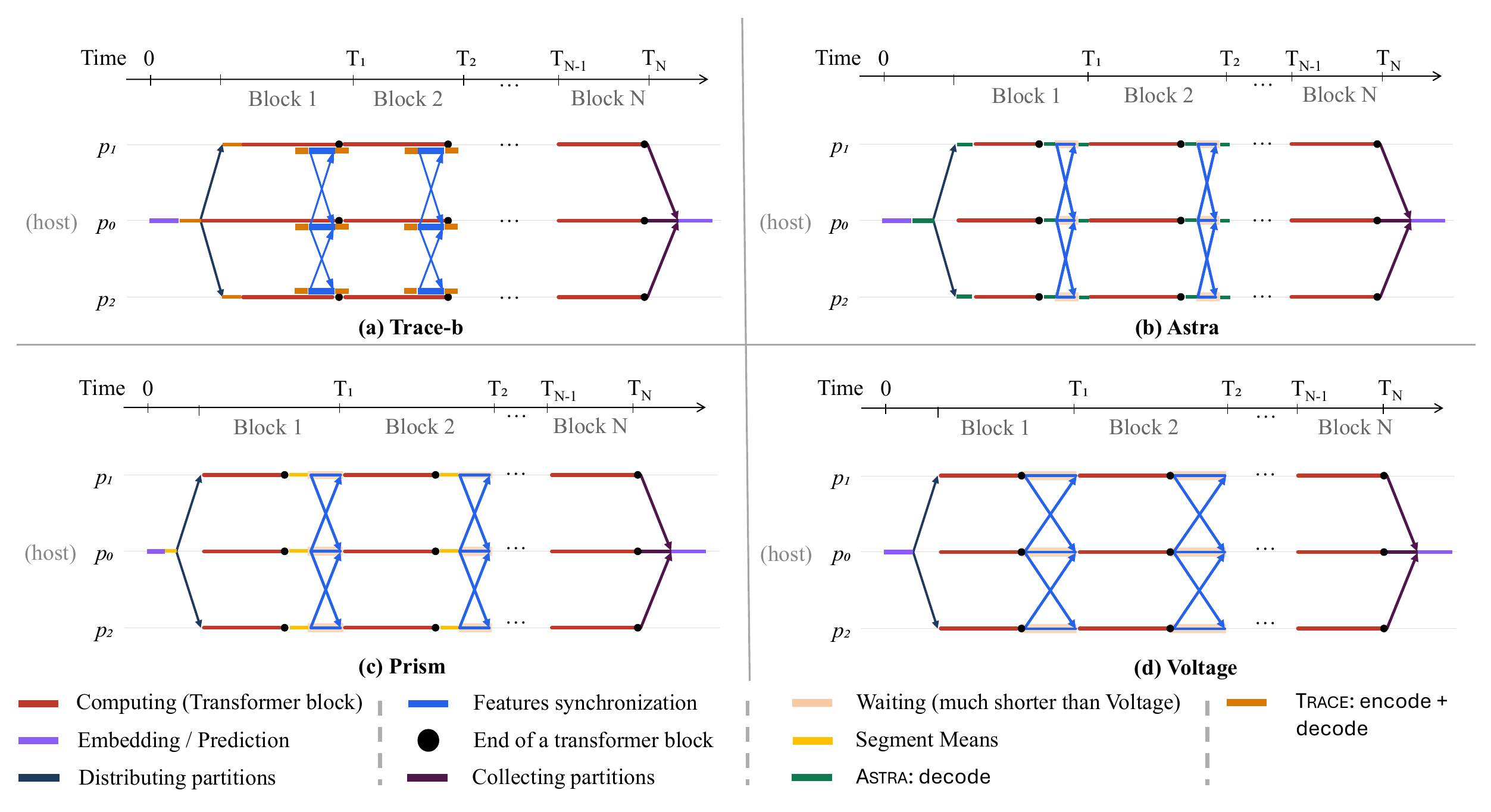}
    \caption{
        \textbf{Compute--communication timing across distributed inference methods} for Transformer blocks across devices $p_0$, $p_1$, $p_2$.
        (a)~\name-b~(ours): index encoding and transmission run in parallel with the local FFN, making communication and computation overlap and saving communication latency.
        (b)~\astra: compact VQ codewords' indices are transmitted, but communication and computation are still non-overlapping.
        (c)~\prism: Segment Means compression reduces communication volume but execution remains sequential.
        (d)~\voltage: computation and communication are fully sequential; all devices idle during full-feature AllGather.
    }
    \label{fig:arrow_diagram}
\end{figure*}

The placement of operation~3 in parallel with operation~2 yields the compute--communication overlap (Fig.~\ref{fig:arrow_diagram}) that distinguishes \name-b from  prior compression methods (Fig.~\ref{fig:methods}): the sender's MLP and projections proceed independently of the index broadcast, and the receiver does no neural-network computation on incoming data, so the only inter-device synchronization point per layer is the broadcast itself. 

\textit{Variant:} The five-operation flow defines \name-b (\emph{buffered}), the default configuration shown in Figs.~\ref{fig:methods}(a) and \ref{fig:arrow_diagram}(a). We also evaluate \name-nb (\emph{non-buffered}), which places VQ before the post-attention residual and uses no input-stage quantizer $\mathrm{VQ}_0$ (Case~1\,A.i in Table~\ref{tab:ablation_placement}). Its first inter-device communication therefore carries the unquantized input partition, while subsequent exchanges transmit codeword indices but reconstruct representations that also depend on runtime residuals. The precomputed K/V buffers are consequently invalid, and the receiver must apply the corresponding transformation chain to the received representations before K/V projection. \name-nb thus preserves runtime residual information and achieves higher accuracy. In contrast, \name-b replaces this computation with buffer lookup, enabling compute--communication overlap. \name-nb is therefore preferable when the application does not have stringent inference latency constraints and the connectivity between edge devices offers high bandwidth, whereas \name-b is preferable when application has stringent inference latency constraints and the network connection is bandwidth-constrained or highly variable.

\begin{table*}[t]
    \centering
    \caption{Summary of notation used throughout this paper.}
    \label{tab:notation}
    \small
    \renewcommand{\arraystretch}{1.05}
    \setlength{\tabcolsep}{4pt}
    \begin{tabular}{@{}p{2.1cm}p{6.35cm}p{1.85cm}p{6.35cm}@{}}
        \toprule
        \textbf{Symbol} & \textbf{Description} & \textbf{Symbol} & \textbf{Description} \\
        \midrule
        \multicolumn{4}{@{}l}{\textit{System and dimensions}} \\
        \midrule
        $P$ & Number of co-located edge devices &
        $L$, $\ell$ & Number of Transformer layers; layer index \\
        $p$, $p'$ & Local device index; peer device index ($p'{\neq}p$) &
        $N$, $N_p$ & Total / per-device sequence length ($N_p{=}N/P$) \\
        $n$ & Token index within a device's local partition &
        $D$ & Model (hidden-state) dimension \\
        $H$, $d$ & Number of attention heads; per-head dimension ($d{=}D/H$) &
        $d_{\mathrm{ff}}$ & Feed-forward (MLP) hidden dimension \\
        \midrule
        \multicolumn{4}{@{}l}{\textit{Transformer activations}} \\
        \midrule
        $\mathbf{X}_p^{(\ell)}$ & Input hidden state of device $p$ at layer $\ell$ &
        $\mathbf{Z}_p^{(\ell)}$ & Post-attention residual activation (quantized by \name) \\
        $\mathbf{Y}_p^{(\ell)}$ & Layer output (post-MLP residual) &
        $\mathbf{Q}_p^{(\ell)},\mathbf{K}_p^{(\ell)},\mathbf{V}_p^{(\ell)}$ & Query, key, value matrices (superscript ``loc'' denotes a device's local-only K/V) \\
        $\mathbf{W}_Q,\mathbf{W}_K,\mathbf{W}_V$ & Query/key/value projection weights &
        $\mathrm{LN}_1,\mathrm{LN}_2$ & Pre-attention / pre-MLP layer normalization \\
        $\mathrm{MHSA}(\cdot)$ & Multi-head self-attention sublayer &
        $\mathrm{MLP}(\cdot)$ & Position-wise feed-forward sublayer \\
        \midrule
        \multicolumn{4}{@{}l}{\textit{Vector quantization and training}} \\
        \midrule
        $\mathcal{C}_\ell$, $K$ & Layer-$\ell$ codebook; codebook size (\# entries) &
        $\mathbf{c}_k^{(\ell)}$ & $k$-th codeword of layer $\ell$ \\
        $\mathrm{VQ}_\ell(\cdot)$, $\mathrm{VQ}_0$ & Per-layer quantizer; input-stage (layer-0) quantizer &
        $k^*_{p,n}$ & Global index of the nearest codeword to token $(p,n)$ \\
        $\tilde{\mathbf{z}}_{p,n}^{(\ell)}$ & Quantized activation (assigned codeword) &
        $\rho$, $\mathcal{M}_p$ & Local-KV drop rate; per-device Bernoulli drop mask \\
        $\alpha$ & EMA codebook decay rate &
        $n_k^{(t)}$, $\mathbf{m}_k^{(t)}$ & EMA count / sum accumulators for codeword $k$ \\
        $\beta$ & Commitment-loss weight &
        $\mathcal{L}$, $\mathcal{L}_{\mathrm{task}}$, $\mathcal{L}_{\mathrm{commit}}^{(\ell)}$ & Total / task / layer-$\ell$ commitment loss \\
        $\mathrm{sg}(\cdot)$ & Stop-gradient operator &
        $\varepsilon$ & Small constant preventing division by zero \\
        $d_q$ & Reduced dimension for the projected EMA variant &
        $\mathbf{W}_{\downarrow},\mathbf{W}_{\uparrow}$ & Down-/up-projection weights for the $d_q$-dim.\ variant \\
        \midrule
        \multicolumn{4}{@{}l}{\textit{Multi-resolution hierarchy and K/V buffers}} \\
        \midrule
        $R$, $m$ & Index of the finest hierarchy level; active hierarchy level ($0{\leq}m{\leq}R$) &
$\mathcal{C}_m$, $K_m$ & Sub-codebook at level $m$; its size \\
        $\boldsymbol{\pi}\in\mathbb{Z}^K$ & Prefix array mapping local $\to$ global codeword index &
        $i_n$, $g_n$ & Transmitted local index; recovered global index ($g_n{=}\boldsymbol{\pi}[i_n]$) \\
        $r = \lceil\log_2 K_m\rceil$ & Bit-width transmitted per token at level $m$ &
        $\mathbf{K}^{\mathsf{vqcb}}_\ell,\mathbf{V}^{\mathsf{vqcb}}_\ell$ & Precomputed K/V lookup buffers ($\in\mathbb{R}^{K\times D}$) \\
        $\mathbf{K}^{\mathrm{vq},(\ell)}_{p'},\mathbf{V}^{\mathrm{vq},(\ell)}_{p'}$ & Buffer-recovered K/V for peer $p'$ at layer $\ell$ &
        $\mathbf{O}_p^{(\ell)}$ & Attention output for device $p$'s local queries at layer $\ell$ \\
        \midrule
        \multicolumn{4}{@{}l}{\textit{Communication}} \\
        \midrule
        $s$ & \prism segment size &
        $\mathcal{BW}$ & Link bandwidth \\
        $t_{\mathrm{comm}}$, $t_{\mathrm{MLP}}$ & Per-layer communication latency; local MLP latency &
        CR & Compression ratio (uncompressed size / transmitted size) \\
        \bottomrule
    \end{tabular}
\end{table*}

\section{Methodology}
\label{sec:method-detail}


This section presents the technical details of \name, including codebook training, post-training K/V calibration, distributed inference, and multi-resolution codebook construction. Table~\ref{tab:notation} summarizes the notation used throughout this section.


\subsection{Codebook Training}
\label{sec:method:training}

\paragraph{VQ insertion}
A vector quantizer $\mathrm{VQ}_\ell$ with a dedicated codebook $\mathcal{C}_\ell = \{\mathbf{c}_k^{(\ell)}\}_{k=1}^{K} \subset \mathbb{R}^D$ is inserted at the post-attention residual of every Transformer layer. For each token $n \in \{1, \ldots, N_p\}$, the nearest codeword is identified by cosine similarity over the full codebook:

\begin{equation}
    \tilde{\mathbf{z}}_{p,n}^{(\ell)}
        = \mathrm{VQ}_\ell\!\left(\mathbf{z}_{p,n}^{(\ell)}\right)
        = \mathbf{c}_{k^*_{p,n}}^{(\ell)},
    \label{eq:vq-insert}
\end{equation}
\begin{equation}
    k^*_{p,n} = \operatorname*{arg\,max}_k\,
        \frac{\mathbf{z}_{p,n}^{(\ell)\top}\,\mathbf{c}_k^{(\ell)}}
             {\|\mathbf{z}_{p,n}^{(\ell)}\|_2\,
              \|\mathbf{c}_k^{(\ell)}\|_2}.
    \label{eq:vq-insert-eucl}
\end{equation}

At inference, nearest codeword search is restricted to the active sub-codebook $\mathcal{C}_m = \{\mathbf{c}_{\boldsymbol{\pi}[0]}^{(\ell)}, \ldots,   \mathbf{c}_{\boldsymbol{\pi}[K_m-1]}^{(\ell)}\}$, and the sender transmits the local position $i_n \in \{0,\ldots,K_m{-}1\}$ within the prefix array $\boldsymbol{\pi}$. The receiver recovers the global codebook index as $g_n = \boldsymbol{\pi}[i_n]$ at all resolution levels and uses it to index the K/V buffers. VQ is placed \emph{after} the residual addition (i.e., on $\mathbf{Z}_p^{(\ell)} = \mathbf{X}_p^{(\ell)} + \mathrm{MHSA}(\cdot)$, not on the raw attention output), so that each codeword index maps deterministically to a unique K/V vector independently of any runtime input. This is the property that makes buffer precomputation exact. The rationale for this placement, and the comparison with alternative designs, are evaluated in Section~\ref{sec:eval:ablation}. An additional input-stage quantizer $\mathrm{VQ}_0$ is applied to the patch or token embeddings before layer~0. It enables the first inter-device exchange to use codebook indices and provides the codewords used to construct $\mathbf{K}^{\mathsf{vqcb}}_0$ and $\mathbf{V}^{\mathsf{vqcb}}_0$. The effect of $\mathrm{VQ}_0$ is included in the same
ablation.
\paragraph{Local-KV drop regularization}
Without explicit regularization, a device can minimize the task loss by relying almost entirely on its own full-precision keys and values while treating cross-device VQ representations as negligible. To prevent this degenerate solution, we introduce \emph{local-KV drop}, i.e., during each finetuning forward pass, a fraction $\rho$ of each device's own key/value tokens are randomly masked before attention is formed, forcing the model to attend across device boundaries. Formally, let $\mathcal{M}_p \sim \mathrm{Bernoulli}(1{-}\rho)^{N_p}$ be a binary mask over the $N_p$ local token positions, resampled at each forward pass. The device-$p$ key matrix entering attention at layer $\ell$ is
\begin{equation}
    \mathbf{K}_p
    = \Bigl[
        \mathbf{K}^{\mathrm{loc}}_p[\mathcal{M}_p]
        \;\Big\|\;
        \bigl\{\mathbf{K}^{\mathrm{vq}}_{p'}\bigr\}_{p' \neq p}
      \Bigr],
    \label{eq:kv-drop}
\end{equation}
with $\mathbf{V}_p$ constructed analogously. We set $\rho{=}0.3$ during finetuning and $\rho{=}0$ at inference, so the full local key/value set is always used at test time.

\paragraph{Codebook update}
Each $\mathrm{VQ}_\ell$ is trained with cosine-similarity assignment and EMA codebook updates~\cite{van2017neural}. Since the same procedure is applied independently at every layer, we suppress the superscript $\ell$ for notational clarity. Let $S_k^{(t)}$ denote the set of post-attention vectors assigned to codeword $\mathbf{c}_k$ in training batch~$t$, across all devices and token positions. The EMA update maintains, for each codeword, a count accumulator $n_k$ and a sum accumulator $\mathbf{m}_k$:

\begin{align}
    n_k^{(t)} &= \alpha\, n_k^{(t-1)}
                  + (1{-}\alpha)\,|S_k^{(t)}|,
                  \label{eq:ema-n} \\
    \mathbf{m}_k^{(t)} &= \alpha\, \mathbf{m}_k^{(t-1)}
                  + (1{-}\alpha)\!\sum_{\mathbf{z} \in S_k^{(t)}}
                  \!\mathbf{z},
                  \label{eq:ema-m} \\
    \mathbf{c}_k^{(t)} &= \mathbf{m}_k^{(t)}
                  \big/ \max\!\left(n_k^{(t)},\,\varepsilon\right),
                  \label{eq:ema-c}
\end{align}
with decay $\alpha{=}0.99$ and $\varepsilon$ a small constant to prevent division by zero. Because the codebook is updated by EMA rather than gradient descent, a commitment loss~\cite{van2017neural} supplies the gradient signal back to the encoder:

\begin{equation}
    \mathcal{L}_{\mathrm{commit}}^{(\ell)}
    = \beta \sum_{p,n}
        \left\|
            \mathbf{z}_{p,n}^{(\ell)}
            - \mathrm{sg}\!\left(\tilde{\mathbf{z}}_{p,n}^{(\ell)}\right)
        \right\|_2^2,
    \label{eq:commit}
\end{equation}
where $\mathrm{sg}(\cdot)$ denotes the stop-gradient operator, $\beta{=}0.1$, and the sum is taken over all devices $p$ and token positions $n$ in the training batch. The commitment loss retains the layer index $\ell$ since it is computed independently per layer. The total training loss combines the task loss with all layer-wise commitment terms:

\begin{equation}
    \mathcal{L} = \mathcal{L}_{\mathrm{task}}
                + \sum_{\ell=0}^{L-1}
                  \mathcal{L}_{\mathrm{commit}}^{(\ell)}.
    \label{eq:total-loss}
\end{equation}

An alternative codebook update strategy projects hidden states to a lower-dimensional space ($d_q{=}256$) before quantization via a learned linear map $\mathbf{W}_{\downarrow} \in \mathbb{R}^{D \times d_q}$, applies EMA-based VQ in $\mathbb{R}^{d_q}$, then reprojects to $\mathbb{R}^D$ via $\mathbf{W}_{\uparrow} \in \mathbb{R}^{d_q \times D}$. This reduces the effective dimensionality of nearest-centroid search and can improve codebook utilization at higher device counts, as the benefit of better-conditioned quantization compounds with more cross-device VQ proxies. This variant is architecturally compatible with K/V buffer precomputation since $\mathbf{W}_{\uparrow}$ is a fixed matrix after training and is absorbed into the calibration pass. \name\ uses the direct EMA update ($d_q{=}D$) as the default to keep all reported results comparable with existing baselines. The dimension-reduction variant is evaluated as an ablation in Section~\ref{sec:eval:ablation} and represents a natural direction for future work.

\paragraph{Initialization and dead-code prevention}
Each codebook $\mathcal{C}_\ell$ is initialized by $k$-means clustering (10 iterations) applied to post-attention+residual activations from the first training batch, placing codewords in high-density regions of the activation space and reducing the risk of early codebook collapse. During training, any codeword whose EMA count $n_k^{(t)}$ falls below a threshold $tr$ is re-initialized by random sampling from the current batch's activations. Codewords with extremely low EMA counts receive no meaningful updates and effectively become inactive, so this mechanism ensures full codebook utilization throughout training.

\subsection{Post-Training K/V Calibration}
\label{sec:method:calibration}

Once training converges and all codebooks are fixed, a single calibration forward pass constructs, for each layer $\ell$, two lookup buffers $\mathbf{K}^{\mathsf{vqcb}}_\ell,  \mathbf{V}^{\mathsf{vqcb}}_\ell \in \mathbb{R}^{K \times D}$ that store (for every codeword) the keys and values that layer $\ell$'s attention would produce if that codeword were its input. Algorithm~\ref{alg:calibration} and the bottom-left panel of Fig.~\ref{fig:trace_architecture} summarize the procedure.

\begin{algorithm}[t]
\caption{Post-training K/V calibration for \name.}
\label{alg:calibration}
\KwIn{Trained model with $L$ layers; codebooks
      $\mathcal{C}_0,\ldots,\mathcal{C}_{L-1}$; frozen weights
      $\{W_K^{(\ell)}, W_V^{(\ell)}, \mathrm{LN}_1^{(\ell)},
        \mathrm{LN}_2^{(\ell)},
        \mathrm{MLP}^{(\ell)}\}_{\ell=0}^{L-1}$.}
\KwOut{Lookup buffers
       $\{\mathbf{K}^{\mathsf{vqcb}}_\ell,
          \mathbf{V}^{\mathsf{vqcb}}_\ell\}_{\ell=0}^{L-1}$.}
\BlankLine
\tcp{Input-stage ($\mathrm{VQ}_0$): produces
     $\mathbf{K}^{\mathsf{vqcb}}_0$,
     $\mathbf{V}^{\mathsf{vqcb}}_0$; no preceding MLP residual.}
$\mathbf{C}_0 \leftarrow [\mathbf{c}_1^{(0)};\,\cdots;\,
    \mathbf{c}_K^{(0)}] \in \mathbb{R}^{K \times D}$\;
$\mathbf{K}^{\mathsf{vqcb}}_0 \leftarrow
    \mathrm{LN}_1^{(0)}(\mathbf{C}_0)\, W_K^{(0)}$\;
$\mathbf{V}^{\mathsf{vqcb}}_0 \leftarrow
    \mathrm{LN}_1^{(0)}(\mathbf{C}_0)\, W_V^{(0)}$\;
\BlankLine
\tcp{Main loop: layer-$\ell$ codewords produce
     $\mathbf{K}^{\mathsf{vqcb}}_{\ell+1}$,
     $\mathbf{V}^{\mathsf{vqcb}}_{\ell+1}$
     for $\ell{+}1 \in \{1,\ldots,L{-}1\}$.}
\For{$\ell = 0,\,1,\,\ldots,\,L{-}2$}{
    $\mathbf{C}_\ell \leftarrow [\mathbf{c}_1^{(\ell)};\,\cdots;\,
        \mathbf{c}_K^{(\ell)}] \in \mathbb{R}^{K \times D}$\;
    $\mathbf{H} \leftarrow \mathbf{C}_\ell
        + \mathrm{MLP}^{(\ell)}\!\left(
          \mathrm{LN}_2^{(\ell)}(\mathbf{C}_\ell)\right)$
        \tcp*{layer-$\ell$ MLP residual}
    $\hat{\mathbf{H}} \leftarrow
        \mathrm{LN}_1^{(\ell+1)}(\mathbf{H})$
        \tcp*{layer-$(\ell{+}1)$ pre-attention LN}
    $\mathbf{K}^{\mathsf{vqcb}}_{\ell+1} \leftarrow
        \hat{\mathbf{H}}\, W_K^{(\ell+1)}$\;
    $\mathbf{V}^{\mathsf{vqcb}}_{\ell+1} \leftarrow
        \hat{\mathbf{H}}\, W_V^{(\ell+1)}$\;
}
\Return $\{\mathbf{K}^{\mathsf{vqcb}}_\ell,\,
           \mathbf{V}^{\mathsf{vqcb}}_\ell\}_{\ell=0}^{L-1}$\;
\end{algorithm}

\paragraph{Data independence and complexity}
The buffers depend solely on the trained codebook entries and the frozen model weights. The dominant cost per layer is the MLP forward pass over $\mathbf{C}_\ell \in \mathbb{R}^{K \times D}$, of order $\mathcal{O}(K D d_{\mathrm{ff}})$ where $d_{\mathrm{ff}}$ is the feed-forward hidden dimension (typically $4D$), together with the K/V projections at $\mathcal{O}(K D^2)$. For $K{=}1024$ and ViT-Base dimensions, calibration completes in seconds on a server-class GPU and need only be performed once.

\paragraph{Storage footprint}
Each layer stores two buffers of shape $K \times D$ at FP16, occupying $4KD$ bytes, making the total across $L$ layers $4KDL$ bytes. For ViT-Base ($K{=}1024$, $D{=}768$, $L{=}12$), this amounts to approximately $36$~MB, substantially smaller than the ${\approx}344$~MB FP32 model weights.

\paragraph{Correctness argument}
At inference, a device receiving cross-device tokens would ordinarily evaluate the chain $\mathrm{LN}_2^{(\ell)} \!\to\! \mathrm{MLP}^{(\ell)}  \!\to\! \text{residual}  \!\to\! \mathrm{LN}_1^{(\ell+1)}  \!\to\! W_K^{(\ell+1)},\,W_V^{(\ell+1)}$ on every incoming feature (Fig.~\ref{fig:trace_architecture} bottom left). Because the sending device constrains each transmitted feature to one of $K$ codewords, this entire chain is replaced by a one-time precomputation over $K$ fixed inputs, i.e., at runtime the receiver performs batched indexed retrieval from $\mathbf{K}^{\mathsf{vqcb}}_{\ell+1}$ and $\mathbf{V}^{\mathsf{vqcb}}_{\ell+1}$, with no matrix multiplication, layer normalization, or nonlinearity applied to the received activations. This determinism holds precisely because VQ is placed after the residual addition (Section~\ref{sec:method:training}). That is, any placement that leaves a runtime-dependent residual term outside the quantizer would break the one-to-one mapping between index and K/V vector, invalidating the buffers.

\subsection{Distributed Inference}
\label{sec:method:inference}
\paragraph{Split-attention}
At layer $\ell$, device $p$ holds its local query, key, and value matrices $\mathbf{Q}_p^{(\ell)},\,\mathbf{K}_p^{\mathrm{loc},(\ell)},\,  \mathbf{V}_p^{\mathrm{loc},(\ell)} \in \mathbb{R}^{N_p \times D}$, computed via Eq.~\eqref{eq:qkv}. For each peer $p' \neq p$, device $p$ has received the cross-device sub-codebook indices $\{i^{(\ell-1)}_{p',n}\}_{n=1}^{N_{p'}}$ transmitted from layer $\ell{-}1$. These are first mapped to global codebook positions $g^{(\ell-1)}_{p',n} = \boldsymbol{\pi}[i^{(\ell-1)}_{p',n}]$, and the corresponding cross-device keys and values are then retrieved by direct buffer lookup:

\begin{equation}
    \mathbf{K}^{\mathrm{vq},(\ell)}_{p'}
        = \mathbf{K}^{\mathsf{vqcb}}_\ell\!\left[
          \{g^{(\ell-1)}_{p',n}\}_n\right], \;\;
    \mathbf{V}^{\mathrm{vq},(\ell)}_{p'}
        = \mathbf{V}^{\mathsf{vqcb}}_\ell\!\left[
          \{g^{(\ell-1)}_{p',n}\}_n\right].
    \label{eq:lookup}
\end{equation}

Since the attention output is invariant to the ordering of key/value pairs~\cite{qazi2025prism}, local and cross-device keys may be concatenated in any order:

\begin{equation}
    \mathbf{K}^{(\ell)} = \mathrm{Concat}\!\left(
        \mathbf{K}_p^{\mathrm{loc},(\ell)},\;
        \left\{\mathbf{K}^{\mathrm{vq},(\ell)}_{p'}\right\}_{p'\neq p}
    \right),
    \label{eq:concat-k}
\end{equation}
and analogously for $\mathbf{V}^{(\ell)}$. Device $p$ then computes attention over its local queries against the global key/value set:

\begin{equation}
    \mathbf{O}_p^{(\ell)} = \mathrm{softmax}\!\left(
        \frac{\mathbf{Q}_p^{(\ell)} \mathbf{K}^{(\ell)\top}}
             {\sqrt{d}}
    \right) \mathbf{V}^{(\ell)}.
    \label{eq:split-attn}
\end{equation}

Concatenating $\{\mathbf{O}_p^{(\ell)}\}_{p=1}^P$ in partition order recovers the full layer output. Because each device computes attention only for its $N_p$ local queries against the full $N$-token key/value set, per-device attention complexity is $\mathcal{O}(N_p N d)$, and the aggregate across all $P$ devices is $\mathcal{O}(N^2 d)$. This is identical to standard sequence-parallel attention. Importantly, unlike other sequence-parallel methods such as \voltage and \astra, where every device must compute full K/V projections for the complete received hidden-state partition, \name\ restricts K/V computation to each device's own local partition only. Cross-device K/V vectors are recovered via direct buffer lookup with no projection matrix multiplications, reducing the per-device K/V projection cost from $\mathcal{O}(N D^2)$ to $\mathcal{O}(N_p D^2)$---a $P$-fold reduction in K/V computation that compounds with the communication savings.

\paragraph{Encoding and Decoding}
For brevity, we omit the device and layer indices and write $i_{p,n}^{(\ell)}=i_n$ and $g_{p,n}^{(\ell)}=g_n$.

\textbf{Encoding} (sender): after cosine-similarity assignment restricted to the active sub-codebook $\mathcal{C}_m$ (Eq.~\eqref{eq:vq-insert}), each token $n$ is assigned a \emph{local} sub-codebook index $i_n \in \{0, \ldots, K_m{-}1\}$, which is its \emph{position} within the prefix array $\boldsymbol{\pi}[0:K_m-1]$, not a direct global codebook index. Concretely, $i_n = j$ means the token was assigned to codeword $\mathbf{c}_{\boldsymbol{\pi}[j]}$; the receiver recovers the global index as $g_n = \boldsymbol{\pi}[i_n]$ at all resolution levels, including the finest ($m{=}R$). An 8-bit metadata header prepended to the frame carries the bit-width $r$, allowing the receiver to decode without out-of-band negotiation and enabling per-session resolution adaptation. Indices produced at layer $\ell$ are used to construct cross-device K/V for layer $\ell{+}1$. Fig.~\ref{fig:trace_architecture}, bottom right, shows an example of the encoded payload. The 8‑bit metadata specifies a bit width of $r=3$, and the subsequent colored boxes represent the 3‑bit indices corresponding to the quantized tokens.

\textbf{Decoding} (receiver): the receiver reads the 8-bit header to recover $r$, then unpacks the $N_{p'}$ local sub-codebook indices $\{i_n\}_{n=1}^{N_{p'}}$ from the $r$-bit fields. Each local index $i_n$ is mapped to a global codebook position $g_n = \boldsymbol{\pi}[i_n]$ via the hierarchy prefix array $\boldsymbol{\pi}$, described in Section~\ref{sec:method:multires}. The corresponding K/V vectors are then retrieved from the precomputed buffers using $g_n$ as in Eq.~\eqref{eq:lookup}. The bottom right panel of Fig.~\ref{fig:trace_architecture} illustrates an example encoded frame and its two-step decoding into buffer lookups.

\paragraph{Compute--communication overlap.}
The index broadcast for partition $p$ and the local MLP forward pass are mutually independent: the indices are derived from $\mathbf{Z}_p^{(\ell)}$, which is fully available before the MLP executes, while the MLP consumes no cross-device data. On systems supporting concurrent execution of network I/O and GPU compute (e.g., CUDA streams with asynchronous communication backends using multi-threading), both operations may therefore proceed simultaneously. The effective per-layer communication latency is thus $\max(t_{\mathrm{comm}}, t_{\mathrm{MLP}}) - t_{\mathrm{MLP}}$, becoming zero whenever $t_{\mathrm{comm}} \leq t_{\mathrm{MLP}}$. This expression represents the ideal overlap enabled by the dependency structure of \name.

\paragraph{Communication volume.}
Table~\ref{tab:comm} compares per-device per-layer communication cost across methods. At resolution bit-width $r$, \name-b transmits $N_p r + 8$ bits per partition per layer, independent of model dimension $D$. At the full codebook ($r{=}10$, $K{=}1024$), \name transmits $10N_p+8$ bits, compared with $32N_pD$ bits for \voltage. The resulting reduction factor is
\begin{equation}
\frac{32N_pD}{10N_p+8},
\end{equation}
which approaches $3.2D$ as $N_p$ increases.

\begin{table}[t]
    \centering
    \caption{Per-device per-layer communication cost (bits transmitted by each device to all peers). $N_p{=}N/P$; $D$: model dimension; $s$: PRISM segment size; $K$: codebook size; $r{=}\lceil\log_2 K_m\rceil$: resolution bit-width.}
    \label{tab:comm}
    \renewcommand{\arraystretch}{1.2}
    \begin{tabular}{@{}lp{3.2cm}c@{}}
        \toprule
        \textbf{Method} & \textbf{Payload (bits)}
                        & \textbf{Depends on $D$?} \\
        \midrule
        \voltage~\cite{10631032}
            & $32 N_p D$
            & Yes \\
        \prism~\cite{qazi2025prism}
            & $32 N_p D / s$
            & Yes \\
        \astra~\cite{liu2025communication}$^\dagger$
            & $N_p\lceil\log_2 K\rceil$
            & No  \\
       \name-b\ (ours)  & $N_p r + 8$ & No \\
\name-nb\ (ours) & $N_p r + 8$ & No \\
        \bottomrule
    \end{tabular}
    \vspace{2pt}
    \begin{minipage}{\columnwidth}
        \footnotesize $^\dagger$\astra reports $N_p\lceil\log_2 K\rceil$ bits theoretically; a na\"{i}ve uint16 implementation transmits $16N_p$ bits per partition. \name\ explicitly bit-packs indices, achieving $N_p r{+}8$ bits including the metadata header.
    \end{minipage}
\end{table}

\paragraph{Layer-0 special case.}
At layer~0, no codeword indices from an upstream layer are available. The input-stage quantizer $\mathrm{VQ}_0$ quantizes the patch or token embeddings directly, and the buffers $\mathbf{K}^{\mathsf{vqcb}}_0$, $\mathbf{V}^{\mathsf{vqcb}}_0$ precomputed in Algorithm~\ref{alg:calibration} are used.


\subsection{Multi-Resolution Codebook}
\label{sec:method:multires}

\paragraph{Hierarchical construction}
After training converges, hierarchical $k$-medoids clustering~\cite{kaufman2009finding} is applied to each layer's trained codebook $\mathcal{C} = \mathcal{C}_R$ to produce a nested hierarchy:

\begin{equation}
    \mathcal{C}_0 \subset \mathcal{C}_1 \subset \cdots
    \subset \mathcal{C}_R = \mathcal{C},
    \qquad
    K_0 < K_1 < \cdots < K_R = K.
    \label{eq:hierarchy}
\end{equation}

We use $(K_0, K_1, K_2, K_3, K_4) = (2, 4, 8, 32, 1024)$. Construction proceeds fine-to-coarse: starting from the full codebook $\mathcal{C}_R$, $k$-medoids is applied to select $K_3{=}32$ representative medoids; those 32 medoids are then clustered to yield $K_2{=}8$, then $K_1{=}4$, and finally $K_0{=}2$ at the coarsest level. Because each level's medoids are selected from the previous (finer) level's medoid set, the strict-subset property $\mathcal{C}_0 \subset \mathcal{C}_1 \subset \cdots \subset \mathcal{C}_R$ is enforced by construction. The resulting levels are stored in coarse-to-fine order in the prefix array $\boldsymbol{\pi}$, defined below.

\paragraph{Hierarchy storage and buffer validity} The hierarchy is stored per-layer as a prefix integer array $\boldsymbol{\pi} \in \mathbb{Z}^K$, where $\boldsymbol{\pi}[i]$ gives the global codebook index corresponding to sub-codebook position $i$; the level boundaries $K_0, K_1, \ldots, K_R$ from Eq.~\eqref{eq:hierarchy} index directly into this array, so that $\boldsymbol{\pi}[0:K_m] = \mathcal{C}_m$ for every level $m$. Decoding a received local index $i$ therefore reduces to two operations: $g \leftarrow \boldsymbol{\pi}[i]$ (prefix lookup), followed by a row fetch from the K/V buffer at row $g$, as described in Section~\ref{sec:method:inference}. Because $\mathcal{C}_m \subset \mathcal{C}_R$ for every $m$, every codeword at any resolution level has a valid precomputed row in $\mathbf{K}^{\mathsf{vqcb}}_\ell$ and $\mathbf{V}^{\mathsf{vqcb}}_\ell$: switching resolution at inference time changes only the nearest-neighbour search domain and the bit-width of the encoded frame; no buffer recomputation or data movement is required.

\paragraph{Runtime resolution selection}
At the start of each inference session, a resolution level $m$ can be selected according to the available link bandwidth and latency budget. The sender restricts cosine-similarity search to $\mathcal{C}_m$ of size $K_m$ and encodes each token's local index in $r = \lceil\log_2 K_m\rceil$ bits. A single trained and calibrated model thereby supports the full family of operating points $\bigl\{(\lceil\log_2 K_m\rceil\;\text{bits/token},\; \mathrm{accuracy}_m)\bigr\}_{m=0}^{R}$ spanning the bitrate--accuracy trade-off frontier without retraining, as reported in Table~\ref{tab:accuracy}.

\section{Experimental Evaluation}
\label{sec:eval}

We evaluate \name\ on image and video classification, comparing against the uncompressed baseline (\voltage), Segment-Means method (\prism), and the strongest prior VQ-based method (\astra). Section~\ref{sec:eval:setup} describes the setup; Sections~\ref{sec:eval:accuracy}--\ref{sec:eval:cr} present accuracy, latency and scalability, communication, compression results; and Section~\ref{sec:eval:ablation} reports four ablations isolating the contribution of each design choice.

\subsection{Experimental Setup}
\label{sec:eval:setup}
\paragraph{Models and datasets} For image classification, ViT-Base/16~\cite{dosovitskiy2020image} is evaluated on ImageNet-1K~\cite{5206848} (single-device baseline accuracy $80.32\%$) and CIFAR-100~\cite{krizhevsky2009learning} ($92.30\%$). For video classification, ViViT-B/16~\cite{arnab2021vivit} is evaluated on UCF-101~\cite{soomro2012ucf101} ($90.77\%$), with scalability measured for $P \in \{2,3,4,6,8,10\}$ devices and evaluated across its multi-resolution sub-codebook hierarchy $K \in \{2,4,8,32,1024\}$.


\paragraph{Training \& Evaluation}
VQ codebooks are finetuned jointly with the pretrained backbone model using EMA codebook updates ($\alpha{=}0.99$), commitment weight $\beta{=}0.1$, local-KV drop $\rho{=}0.3$, and $K{=}1024$ codewords initialized by $k$-means on the first batch. The multi-resolution hierarchy is built post-training by $k$-medoids. Compute latency is measured on a CPU configured with two cores, and communication latency is computed at a range of network bandwidth $\mathcal{BW} = \{1,2,4,8,10,20\}$~Mbps. All latency figures report effective per-device latency averaged over fifty runs ($\pm5$~ms standard deviation).

\subsection{Compression Ratio}
\label{sec:eval:cr}

\begin{figure}[t]
\centering
\includegraphics[width=\columnwidth]{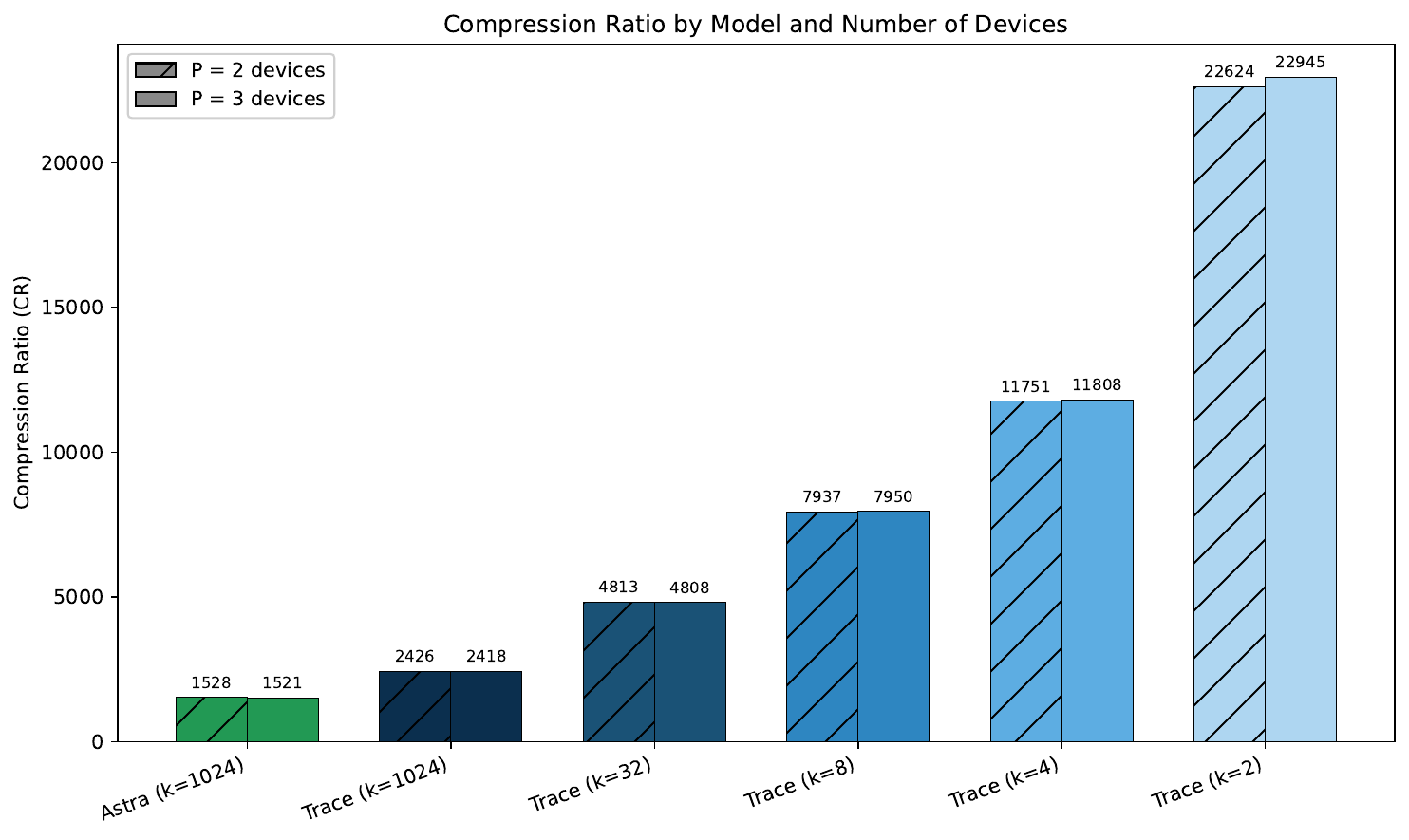}
\caption{Compression ratio for ViT-Base/16 at $P=2$ and $P=3$. \name{} ($K=1024$) achieves $2{,}426\times$ at $P=2$, a $1.59\times$ improvement over \astra{} ($1{,}528\times$) at the same codebook size. Coarser levels yield near-proportional gains, reaching $22{,}624\times$ at $K=2$.}
\label{fig:vit_cr}
\end{figure}

\begin{figure*}[t]
\centering

\subfloat[$P=2$\label{fig:vivit_cr_p2}]{
    \includegraphics[width=0.31\textwidth]{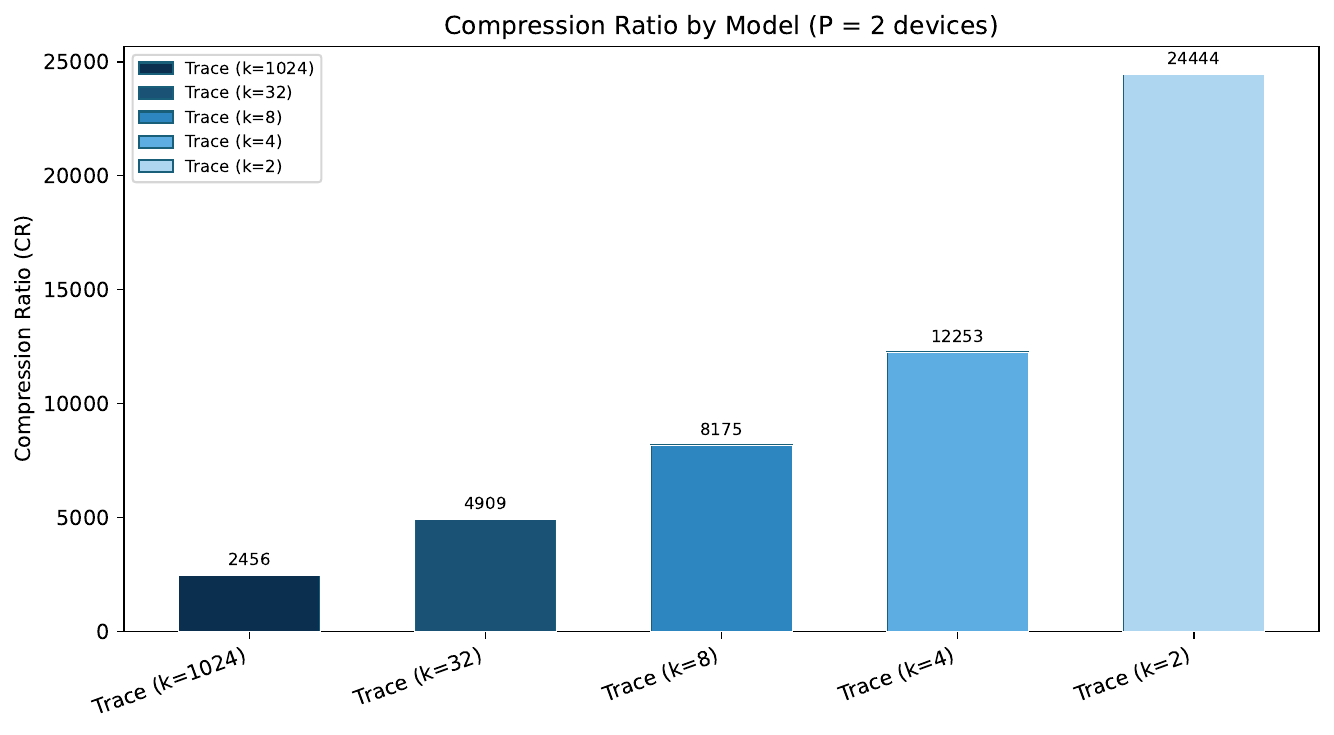}
}
\hfill
\subfloat[$P=4$\label{fig:vivit_cr_p4}]{
    \includegraphics[width=0.31\textwidth]{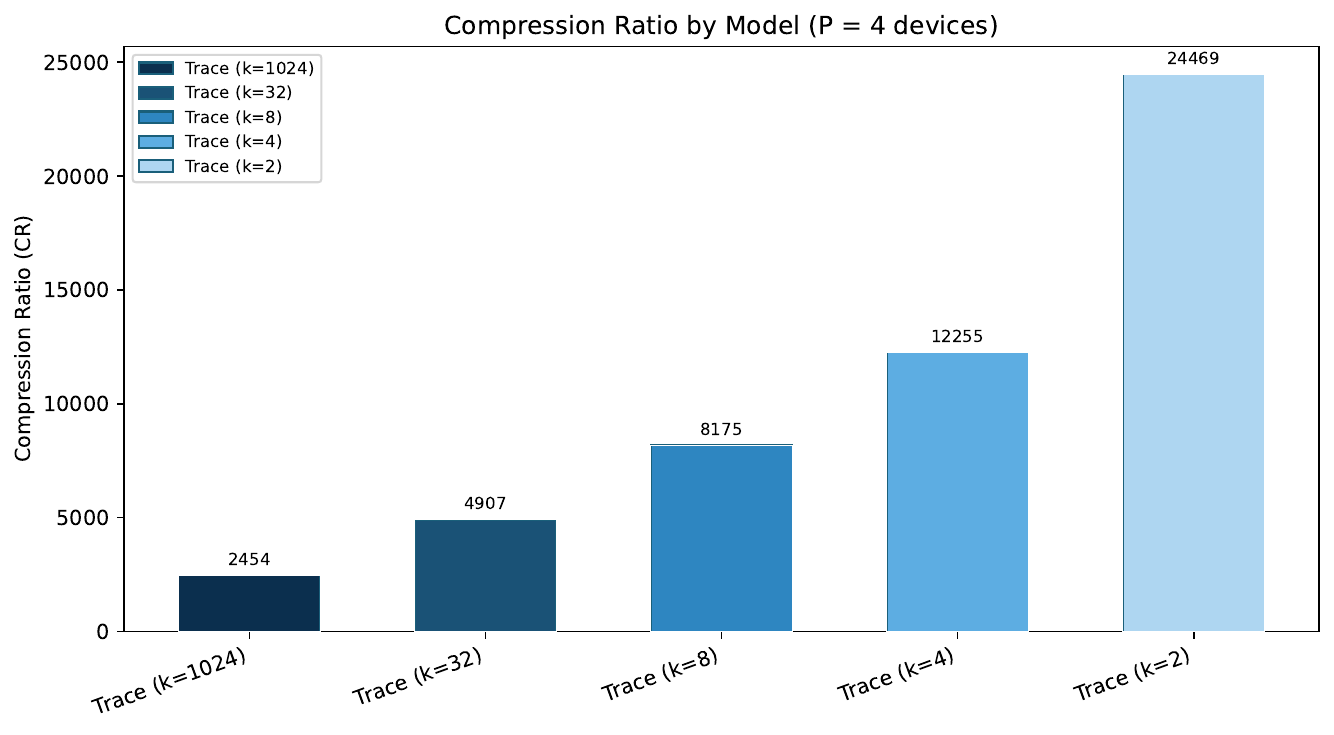}
}
\hfill
\subfloat[$P=10$\label{fig:vivit_cr_p10}]{
    \includegraphics[width=0.31\textwidth]{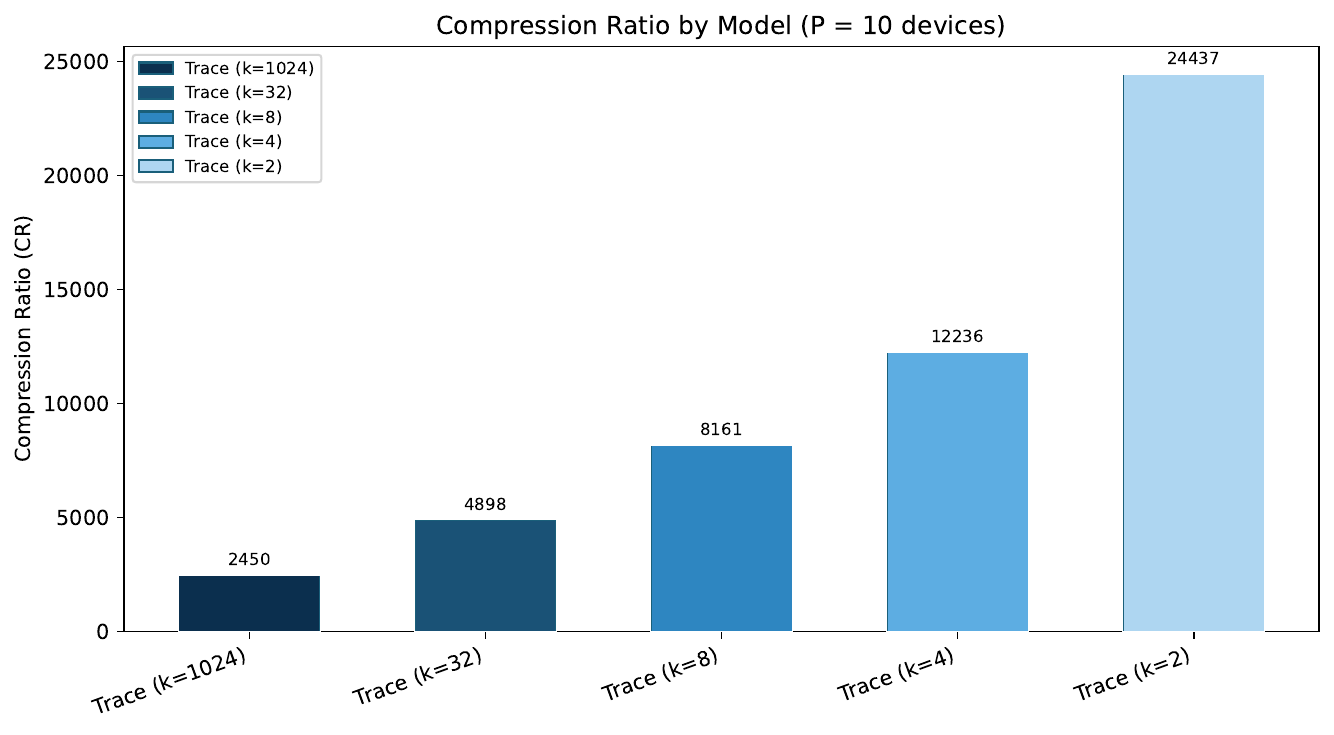}
}

\caption{Compression ratio for ViViT-B/16 on UCF-101 across $P \in \{2,4,10\}$ devices. The compression ratio at each resolution level varies by less than $0.5\%$ across all device counts, confirming that \name{}'s compression mechanism is invariant to the number of participating devices.} \label{fig:vivit_cr}

\end{figure*}

Fig.~\ref{fig:vit_cr} reports the compression ratio (CR), defined as the ratio of uncompressed intermediate features to compressed transmitted data. For ViT at $P{=}2$, \name\ ($K{=}1024$) achieves $2{,}426\times$ compression versus \astra's $1{,}528\times$ at the same codebook size, which is $1.59\times$ improvement that stems from \name\ transmitting $\lceil\log_2 K\rceil$ bits per token independently of model's feature dimension. Successive resolution levels yield near-proportional CR gains: $4{,}813\times$ ($K{=}32$), $7{,}937\times$ ($K{=}8$), and $22{,}624\times$ ($K{=}2$). Fig.~\ref{fig:vivit_cr} confirms that these ratios are essentially invariant to device count for ViViT: across $P \in \{2,4,10\}$, the CR for each resolution level varies by less than $0.5\%$, indicating that the compression mechanism does not degrade as more devices participate, as long as the number of tokens is sufficient large in each partition.

\subsection{Inference Latency} \label{sec:eval:latency}

\begin{figure*}[t]
\centering

\subfloat[$P=2$\label{fig:latency_p2}]{
    \includegraphics[width=0.48\textwidth]{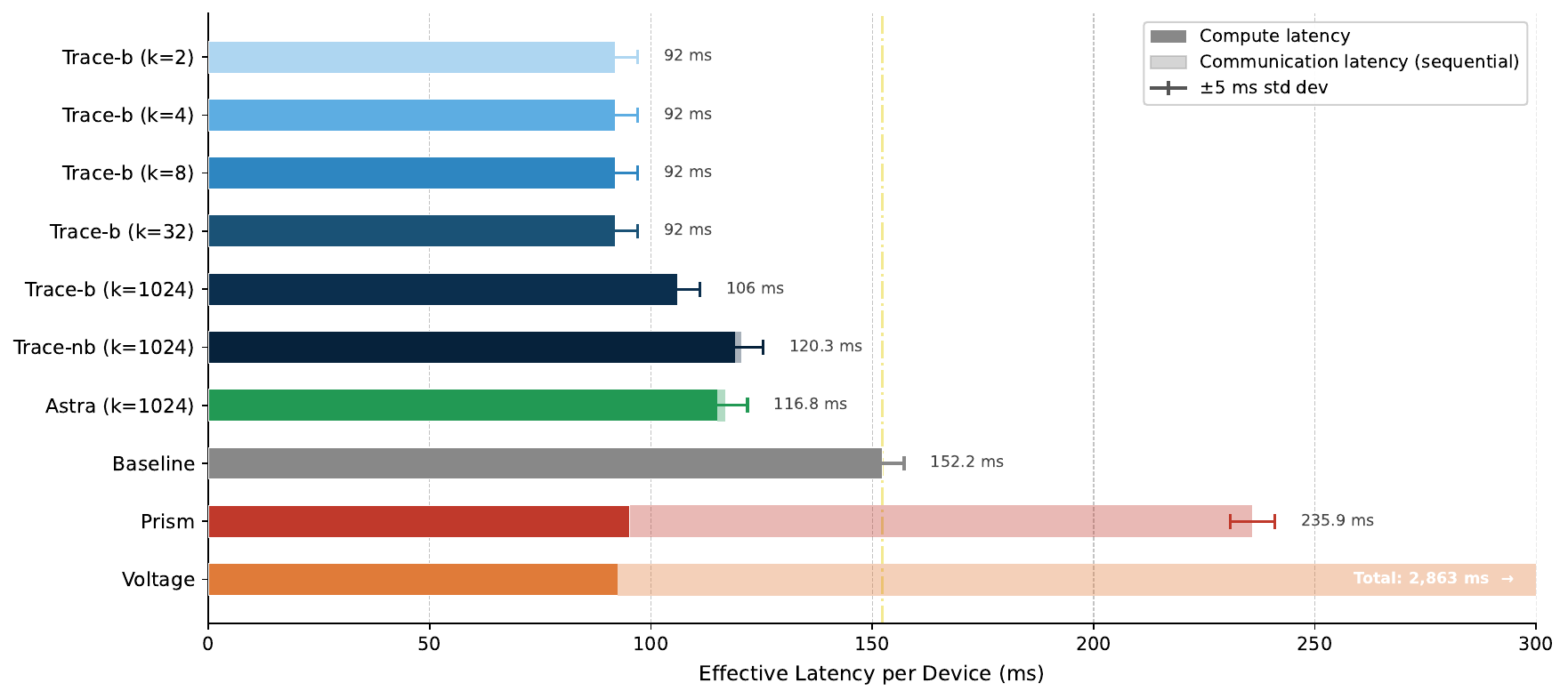}
}
\hfill
\subfloat[$P=3$\label{fig:latency_p3}]{
    \includegraphics[width=0.48\textwidth]{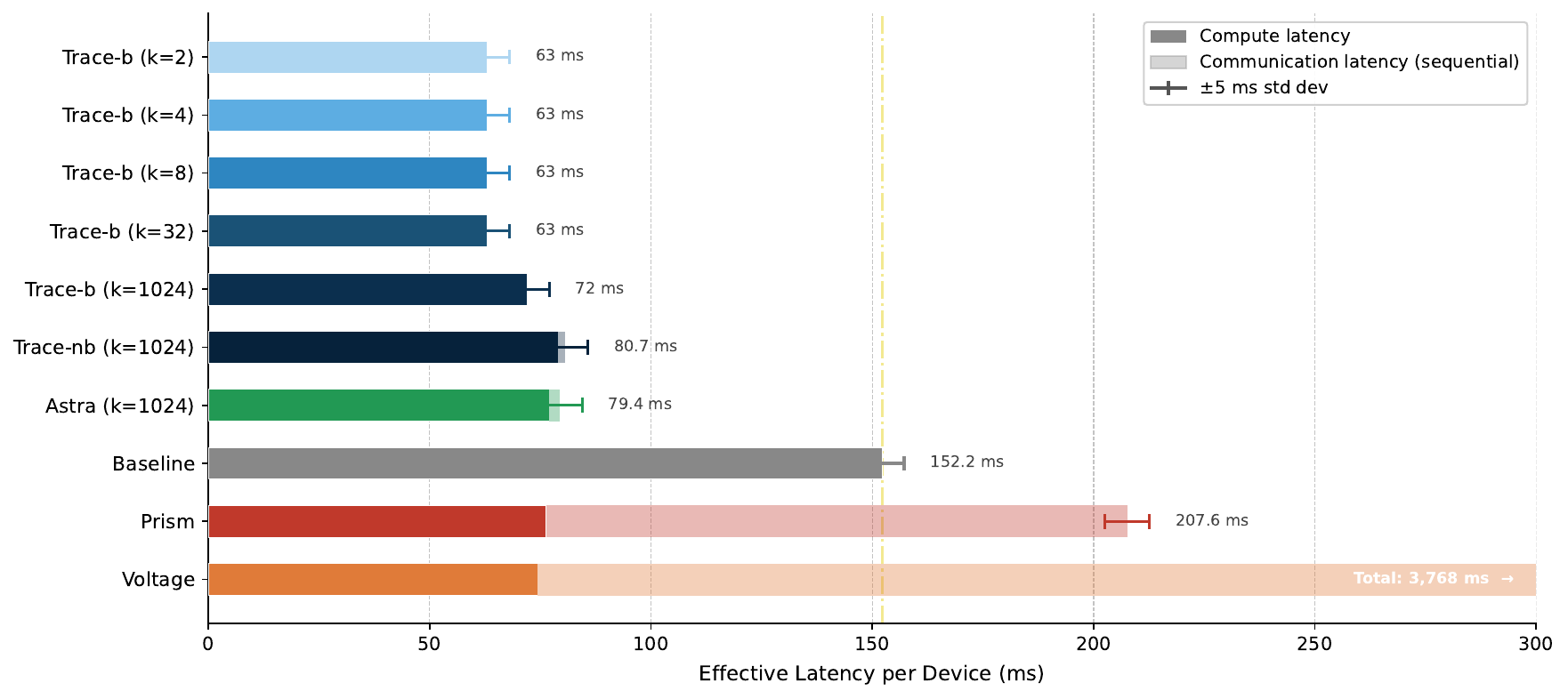}
}

\caption{Effective per-device latency (ms) at 10~Mbps on ViT-Base/16. Dark bars show compute latency; light extensions indicate communication latency when executed sequentially. \name-b at $K\leq32$ reaches the single-device compute floor (92~ms at $P=2$, 63~ms at $P=3$).} \label{fig:latency}

\end{figure*}

\begin{figure}[t]
\centering

    \includegraphics[width=0.48\textwidth]{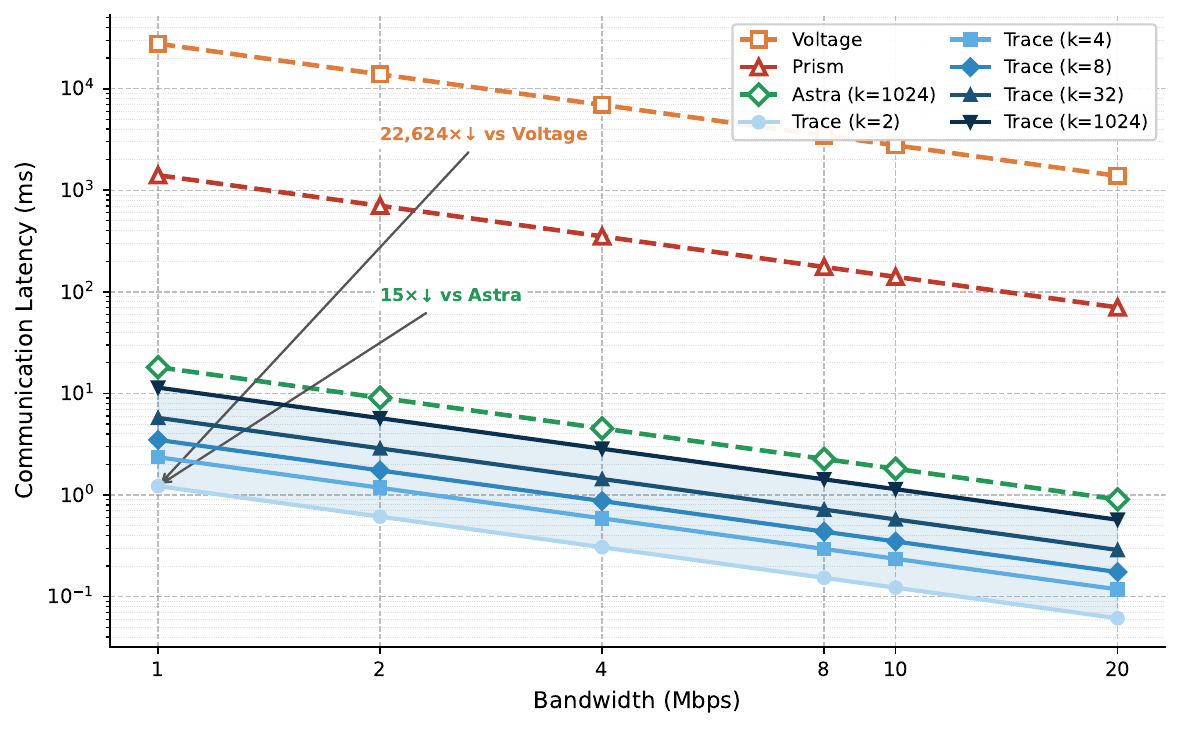}


\caption{Communication latency (ms) versus link bandwidth (Mbps) on ViT-Base/16 (log scale). \name{} ($K=2$) achieves $22{,}624\times$ lower communication latency than \voltage{} and $15\times$ lower than \astra{} at $P=2$. The \name{} curves indicate that its communication advantage is maintained consistently across the evaluated 1--20~Mbps bandwidth range.}
\label{fig:comm}

\end{figure}

Figs.~\ref{fig:latency_p2} and~\ref{fig:latency_p3} show the effective per-device inference latency of ViT-Base/16 at $10$~Mbps, decomposed into computation and communication, while Table~\ref{tab:latency_breakdown} reports the corresponding measurements. Fig.~\ref{fig:comm} further evaluates communication latency over the $1$--$20$~Mbps bandwidth range. Together, the results show that \name-b reduces both communication overhead and receiver-side computation, resulting in the lowest effective latency among the evaluated methods.


At $P{=}2$, \name-b requires only $0.12$--$1.14$~ms of communication by transmitting compact codeword indices, depending on the resolution $m$. Unlike \astra, \name-b also avoids receiver-side K/V projections by retrieving cross-device K/V vectors from precomputed buffers. Consequently, its computation latency decreases to $92$~ms at $P{=}2$ and $63$~ms at $P{=}3$ for $K{\leq}32$. At $K{=}1024$, the larger nearest-neighbor search increases computation modestly to $106$~ms and $72$~ms, respectively. In comparison, \astra requires $1.81$~ms of communication, while \prism reduces communication to $140.63$~ms through Segment-Means compression, although its communication time still exceeds its computation time at $10$~Mbps. The uncompressed \voltage baseline incurs $2{,}770.31$~ms of communication at $10$~Mbps, approximately $30\times$ its $92.5$~ms computation latency, making communication the dominant contributor to end-to-end latency.

The communication advantage is maintained across the evaluated bandwidth range in Fig.~\ref{fig:comm}. As bandwidth decreases, the latency of all methods increases, but the effect is most severe for methods that transmit high-dimensional features. At $P{=}2$, \name-b with $K{=}2$ achieves up to $22{,}624\times$ lower communication latency than \voltage and $15\times$ lower latency than \astra. 
Even at $1$~Mbps, \name-b keeps communication in the millisecond range, whereas \voltage incurs delays on the order of tens of seconds. These results demonstrate that the communication advantage of \name-b is consistent across the evaluated bandwidth range.

Finally, because communication is fully overlapped with local computation (Section~\ref{sec:method:inference}), \name's effective latency at $10$~Mbps equals its computation latency whenever $t_{\mathrm{comm}} \leq t_{\mathrm{MLP}}$. At matched codebook size ($K{=}1024$), eliminating the receiver-side K/V projection alone reduces effective latency from $116.8$ to $106.0$~ms relative to \astra at $P{=}2$ ($1.10\times$), and from $79.4$ to $72.0$~ms at $P{=}3$ ($1.10\times$). \name's hierarchical codebook further lets the same deployed model drop to a coarser resolution ($K{=}32$) with no retraining (a capability \astra's fixed-bitrate design cannot match) bringing effective latency to $92$~ms and $63$~ms, or $1.27\times$ and $1.26\times$ over \astra. 

\subsection{Accuracy versus Compression}
\label{sec:eval:accuracy}


\begin{table}[t]
    \centering
    \caption{Top-1 accuracy (\%) on ImageNet and CIFAR-100
             (ViT-Base/16). Single-device baseline: $80.32\%$
             (ImageNet), $92.30\%$ (CIFAR-100).}
    \label{tab:accuracy}
    \renewcommand{\arraystretch}{1.15}
    \setlength{\tabcolsep}{4.5pt}
    \begin{tabular}{@{}lcccc@{}}
        \toprule
        & \multicolumn{2}{c}{\textbf{ImageNet}}
        & \multicolumn{2}{c}{\textbf{CIFAR-100}} \\
        \cmidrule(lr){2-3}\cmidrule(lr){4-5}
        \textbf{Method} & $P{=}2$ & $P{=}3$ & $P{=}2$ & $P{=}3$ \\
        \midrule
        \voltage~\cite{10631032}      & 80.32 & 80.32 & 92.30 & 92.30 \\
        \astra\,($K{=}1024$)~\cite{liu2025communication}
                                      & 78.45 & 77.14 & 89.34 & 88.12 \\
        \prism~\cite{qazi2025prism}   & 79.66 & 79.47 & 90.10 & 89.69 \\
        \midrule
        \name-nb\,($K{=}1024$)           &\textbf{80.23} & \textbf{80.11} & \textbf{91.23} & \textbf{90.58} \\
        \name-b\,($K{=}1024$)           & 79.90 & 79.70 & 90.55 & 89.90 \\
        \name-b\,($K{=}32$)             & 79.60 & 79.01 & 90.52 & 88.54 \\
        \name-b\,($K{=}8$)              & 79.51 & 78.60 & 89.89 & 87.70 \\
        \name-b\,($K{=}4$)              & 79.46 & 77.29 & 89.18 & 85.81 \\
        \name-b\,($K{=}2$)              & 78.84 & 76.85 & 84.91 & 83.33 \\
        \bottomrule
    \end{tabular}
\end{table}

\begin{figure*}[t]
\centering

\subfloat[ImageNet-1K\label{fig:imagenet_acc}]{
    \includegraphics[width=0.48\textwidth]{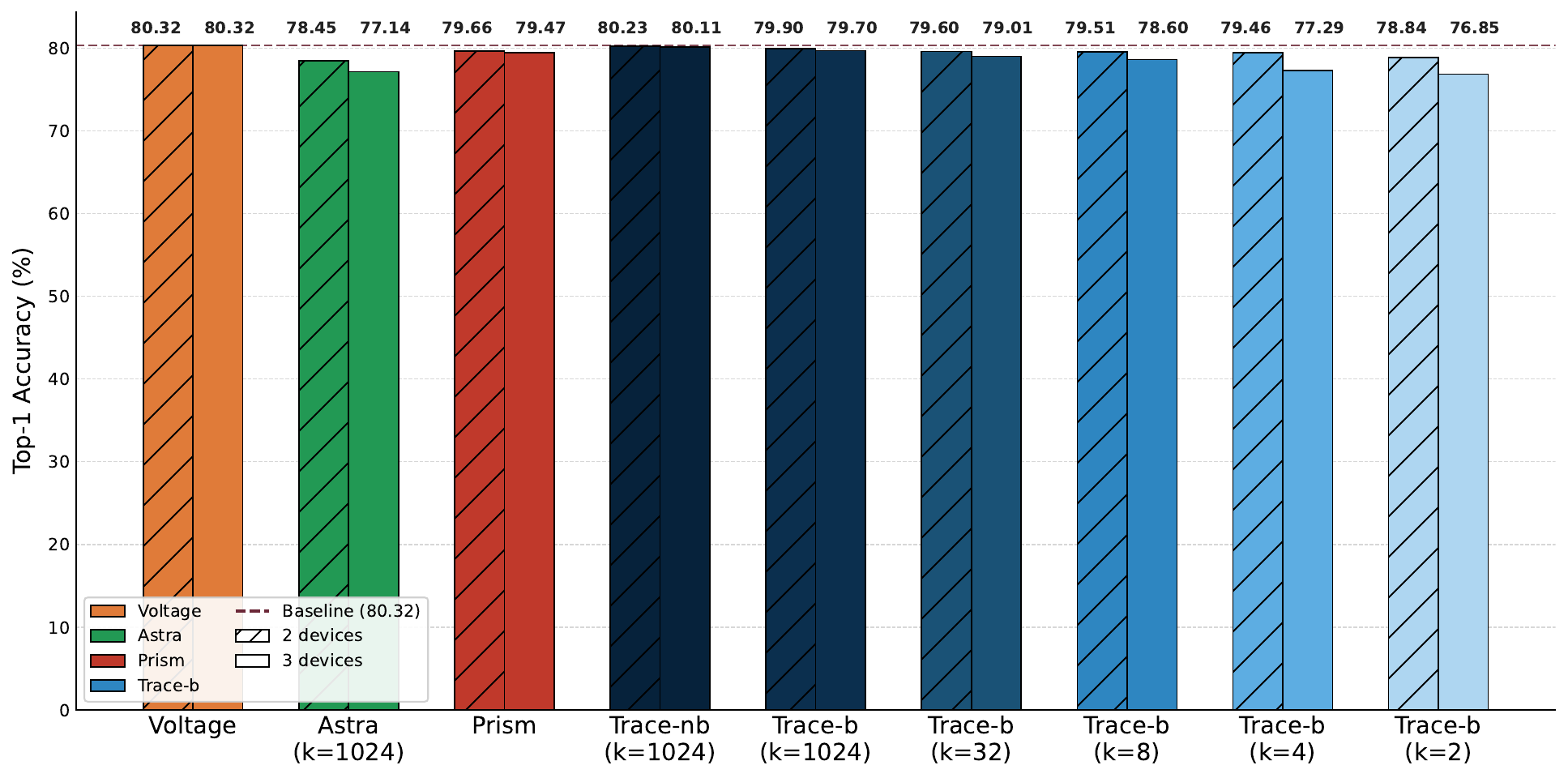}
}
\hfill
\subfloat[CIFAR-100\label{fig:cifar_acc}]{
    \includegraphics[width=0.48\textwidth]{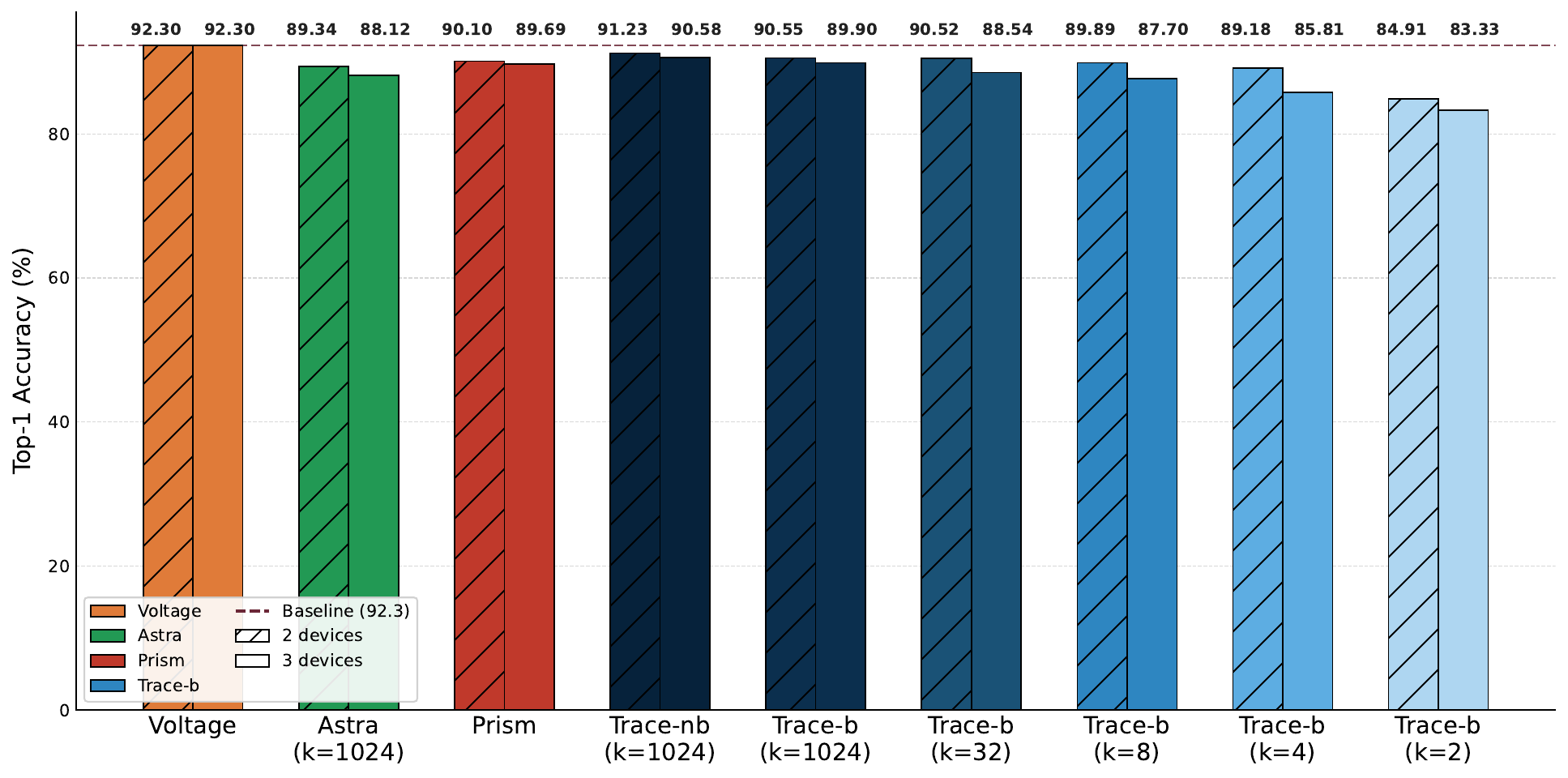}
}


\caption{Top-1 accuracy on ImageNet-1K and CIFAR-100 using ViT-Base/16 with $P=2$ and $P=3$ devices. The single-device baselines are $80.32\%$ on ImageNet-1K and $92.30\%$ on CIFAR-100. At $K=1024$, \name{} achieves the highest accuracy among the evaluated compression methods in all four settings.}

\label{fig:accuracy}

\end{figure*}

Table~\ref{tab:accuracy} and Figs.~\ref{fig:imagenet_acc}--\ref{fig:cifar_acc} show the accuracy--compression trade-off on ImageNet-1K and CIFAR-100. At $K{=}1024$, both \name variants outperform the prior compression baselines, \astra and \prism, across all four settings. \name-nb consistently achieves higher accuracy, for example, $80.23\%$ on ImageNet-1K and $91.23\%$ on CIFAR-100 at $P{=}2$, which is 1.78 and 1.89 percentage point higher than \astra. \name-b still achieves 1.45 and 1.21 percentage point higher accuracy than \astra at $P{=}2$. The trade-off is that \name-nb preserves higher accuracy, while \name-b trades a modest accuracy loss for exact buffer-based K/V reconstruction and compute--communication overlap, achieving inference acceleration. This difference highlights the accuracy--efficiency trade-off introduced by the post-residual placement required for buffer precomputation.

\begin{figure}[t]
    \centering
    \includegraphics[width=\columnwidth]{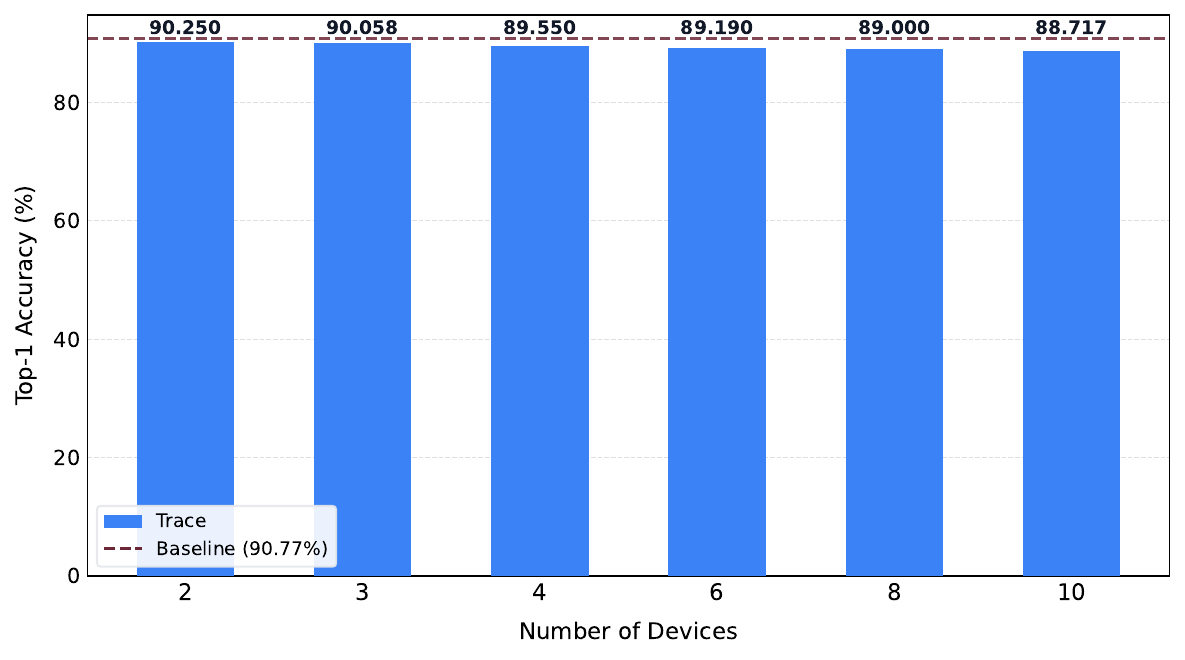}
    \caption{Top-1 accuracy on UCF-101 (ViViT-B/16, $K{=}1024$) as the number of devices scales from $2$ to $10$. \name\ degrades by only $1.53$ percentage points from $P{=}2$ ($90.25\%$) to $P{=}10$ ($88.72\%$) relative to the single-device baseline ($90.77\%$, dashed line).}
    \label{fig:vivit_scalability}
\end{figure}

The hierarchy provides additional compression at an accuracy-dependent trade-off. On ImageNet-1K at $P{=}2$, reducing the effective codebook size from $K{=}1024$ to $K{=}4$ decreases accuracy by only $0.44$ percentage points, while the most aggressive $K{=}2$ level retains $78.84\%$ accuracy using one bit per token. Similarly on UCF-101, Fig.~\ref{fig:vivit_scalability} shows that \name with $K{=}1024$ achieves $90.25\%$ accuracy at $P{=}2$ and $88.72\%$ at $P{=}10$. This runtime-selectable trade-off is particularly relevant under low or time-varying bandwidth, where the system can select a coarser resolution to reduce communication latency while retaining usable prediction accuracy, without retraining the deployed model. This flexibility is valuable for latency-sensitive edge applications, where meeting a latency budget may be more important than maximizing accuracy at a fixed operating point.

\begin{table}[t]
    \centering
    \caption{Per-device latency breakdown (ms) at $10$~Mbps on ViT-Base/16 (2-core CPU).}
    \label{tab:latency_breakdown}
    \renewcommand{\arraystretch}{1.15}
    \setlength{\tabcolsep}{4.5pt}
    \begin{tabular}{@{}lcccc@{}}
        \toprule
        \textbf{Method} & \textbf{$P$}
            & \textbf{Compute} & \textbf{Comm.}
            & \textbf{Effective}$^\dagger$ \\
        \midrule
        \voltage~\cite{10631032}          & 2 & 92.5  & 2770.31 & 2862.8 \\
        \prism\,($s{=}10$)~\cite{qazi2025prism}
                                          & 2 & 95.2  & 140.63  & 235.9 \\
        \astra\,($K{=}1024$)~\cite{liu2025communication}
                                          & 2 & 115.0 & 1.81    & 116.8 \\
        \name-nb\,($K{=}1024$)               & 2 & 119.0 & 1.34    & 120.3 \\
        \name-b\,($K{=}1024$)               & 2 & 106.0 & 1.14    & \textbf{106.0} \\
        \name-b\,($K{=}32$)                 & 2 & 92.0  & 0.58    & \textbf{92.0} \\
        \name-b\,($K{=}2$)                  & 2 & 92.0  & 0.12    & \textbf{92.0} \\
        \midrule
        \voltage~\cite{10631032}          & 3 & 74.4  & 3693.75 & 3768.1 \\
        \prism\,($s{=}10$)~\cite{qazi2025prism}
                                          & 3 & 76.3  & 131.25  & 207.6 \\
        \astra\,($K{=}1024$)~\cite{liu2025communication}
                                          & 3 & 77.0  & 2.43    & 79.4 \\
        \name-nb\,($K{=}1024$)               & 3 & 79.0  & 1.68    & 80.7 \\
        \name-b\,($K{=}1024$)               & 3 & 72.0  & 1.53    & \textbf{72.0} \\
        \name-b\,($K{=}32$)                 & 3 & 63.0  & 0.77    & \textbf{63.0} \\
        \name-b\,($K{=}2$)                  & 3 & 63.0  & 0.16    & \textbf{63.0} \\
        \bottomrule
    \end{tabular}
    \\[3pt]

        \begin{minipage}{\columnwidth}
        \footnotesize $^\dagger$\emph{Effective} latency accounts for compute--communication overlap. For \name-b, communication is overlapped with local MLP execution; for the baselines, communication and computation are sequential.
    \end{minipage}

\end{table}

\subsection{Ablation Studies}
\label{sec:eval:ablation}

We isolate the contribution of four design choices: the precomputed K/V buffers, the VQ placement, the hierarchy construction metric, and the codebook update strategy.

\paragraph{Precomputed K/V buffers}
\label{sec:eval:ablation:buffer} Table~\ref{tab:ablation_buffer} quantifies the per-device compute savings from buffer precomputation on ViViT. Comparing \name-b against an otherwise identical configuration in which the buffers are disabled and the receiver reconstructs cross-device K/V through the full $\mathrm{LN}\!\to\!\mathrm{MLP}\!\to\!\mathrm{residual}\!\to\!\mathrm{K/V}$ chain. This configuration retains \name-b's post-residual VQ placement and is therefore distinct from \name-nb, which additionally moves VQ before the residual.
The buffers reduce per-device compute latency by $15.5$--$32.3\%$.

\begin{table}[t]
    \centering
    \caption{Effect of precomputed K/V buffers on per-device compute latency (ms) for ViViT ($K{=}1024$). ``Without'' reconstructs cross-device K/V via the full projection chain; ``With'' uses buffer lookup.}
    \label{tab:ablation_buffer}
    \renewcommand{\arraystretch}{1.15}
    \begin{tabular}{@{}cccc@{}}
        \toprule
        \textbf{Devices} & \textbf{Without buffers}
                         & \textbf{With buffers}
                         & \textbf{Gain (\%)} \\
        \midrule
        2  & 3441 & 2329 & 32.3 \\
        3  & 2281 & 1596 & 30.0 \\
        4  & 1766 & 1320 & 25.2 \\
        6  & 1213 &  952 & 21.5 \\
        8  &  917 &  719 & 21.6 \\
        10 &  742 &  627 & 15.5 \\
        \bottomrule
    \end{tabular}
\end{table}

\paragraph{VQ placement} 
\label{sec:eval:ablation:placement}

\name-b places VQ after the post-attention residual and uses $\mathrm{VQ}_0$ for the first block (Case~2\,B in Table~\ref{tab:ablation_placement}). Consequently, each transmitted activation is represented entirely by a codeword index, with no runtime-dependent residual component. This establishes an exact index-to-K/V mapping, allowing non-local K/V vectors to be retrieved from the precomputed buffers. The use of $\mathrm{VQ}_0$ also ensures that the first inter-device exchange is performed in the codebook domain. This design achieves top-1 accuracies of $90.51\%$ and $89.90\%$ for $P{=}2$ and $P{=}3$, respectively.

We compare this design against VQ placements before and after the residual. Case~2\,A retains the post-residual placement but computes the first-block K/V from the unquantized input, leaving the first exchange uncompressed. In the Case~1 variants, VQ is applied before the residual, so the transmitted representation carries a runtime-dependent residual and exact codebook-indexed retrieval from K/V buffers is no longer possible. Case~1\,A.i attains the highest accuracy of all placements and is adopted as the second variant \name-nb (Section~\ref{sec:system-dataflow}): relative to \name-b it gains $0.72$ and $0.68$ percentage points at $P{=}2$ and $P{=}3$, at the cost of receiver-side reconstruction and the loss of compute--communication overlap. Cases~1\,A.ii, 1\,A.iii and~1\,B lose accuracy relative to 1\,A.i without recovering the deterministic mapping, and are therefore dominated by both variants.

\begin{table}[t]
    \centering
    \caption{VQ placement ablation on CIFAR-100 (Accuracy \%) using
    ViT-Base/16. \name uses post-residual VQ with
    $\mathrm{VQ}_0$ for the first-block K/V.}
    \label{tab:ablation_placement}
    \renewcommand{\arraystretch}{1.15}
    \setlength{\tabcolsep}{4pt}
    \begin{tabular}{@{}lm{5.8cm}cc@{}}
        \toprule
        \textbf{Case} & \textbf{Quantized-copy treatment}
                      & $P{=}2$ & $P{=}3$ \\
        \midrule
                1\,A.i & $\mathbf{X}_p^{(0)}$ residual at block~1;
                 $\tilde{\mathbf{Z}}_p^{(\ell-1)}$ thereafter (\name-nb)
               & 91.23 & 90.58 \\
        2\,B & Post-residual VQ; $\mathrm{VQ}_0$
               K/V at block~1 (\name-b)
             & 90.51 & 89.90 \\
        \midrule
        2\,A & Post-residual VQ; full $\mathbf{X}$
               K/V at block~1
             & 90.52 & 88.61 \\
        1\,A.ii & No residual at block~1;
                  $\tilde{\mathbf{Z}}_p^{(\ell-1)}$ thereafter
                & 90.78 & 89.03 \\
        1\,A.iii & No residual in any block
                  & 90.51 & 88.59 \\
        1\,B & $\mathrm{VQ}_0$ as the block-1 residual and K/V
             & 90.90 & 89.56 \\
        \bottomrule
    \end{tabular}
\end{table}

\paragraph{Hierarchy construction metric}
Table~\ref{tab:ablation_hierarchy} compares three dissimilarity measures for the $k$-medoids hierarchy: Euclidean distance, cosine distance, and codeword usage frequency $n_k^{(t)}$. All three coincide at the full codebook ($m{=}R$); differences emerge only at coarser levels. Euclidean and cosine perform comparably, with neither dominating across all settings, while frequency-based clustering underperforms markedly at $P{=}3$ and coarse levels, hence grouping codewords by popularity rather than geometric proximity yields medoids that are less representative of their clusters. We adopt Euclidean $k$-medoids as the default.

\begin{table}[t]
    \centering
    \caption{Hierarchy construction metric ablation: top-1
             accuracy (\%) on CIFAR-100 at each resolution
             level $m$ (level $4$ = full codebook, $K{=}1024$).}
    \label{tab:ablation_hierarchy}
    \renewcommand{\arraystretch}{1.15}
    \setlength{\tabcolsep}{4pt}
    \begin{tabular}{@{}clccccc@{}}
        \toprule
        & & \multicolumn{5}{c}{\textbf{Resolution level $m$}} \\
        \cmidrule(lr){3-7}
        \textbf{$P$} & \textbf{Metric}
            & 0 & 1 & 2 & 3 & 4 \\
        \midrule
        \multirow{3}{*}{2}
          & Euclidean & 84.91 & 89.18 & \textbf{89.89} & \textbf{90.52} & 90.55 \\
          & Cosine    & 83.30 & 88.40 & 89.53 & 90.32 & 90.55 \\
          & Frequency & \textbf{85.56} & \textbf{89.29} & 89.73 & 89.72 & 90.55 \\
        \midrule
        \multirow{3}{*}{3}
          & Euclidean & 83.33 & 85.81 & \textbf{87.70} & 88.54 & 89.90 \\
          & Cosine    & \textbf{82.59} & \textbf{85.30} & 86.28 & \textbf{87.60} & 88.90 \\
          & Frequency & 80.04 & 80.62 & 84.40 & 85.04 & 88.90 \\
        \bottomrule
    \end{tabular}
\end{table}

\paragraph{Codebook update strategy}
Table~\ref{tab:ablation_update} compares gradient-based codebook updates, EMA (\name's default), and EMA applied in a dimension-reduced subspace ($d_q{=}256$). EMA outperforms gradient-based updates at every device count, with the gap widening from $0.04$ points at $P{=}2$ to $0.56$ at $P{=}3$, consistent with the known instability of gradient-based codebook training under multiple VQ bottlenecks. Applying EMA in a $256$-dimensional projected space improves accuracy by a further $0.28$--$0.30$ points, with the benefit growing with $P$: better-conditioned nearest-centroid search in lower dimensions becomes increasingly valuable as cross-device proxies multiply. This variant is compatible with buffer precomputation (the up-projection is absorbed into calibration) and is a promising direction for future work; \name\ uses the direct EMA update as the default to keep all reported results comparable.

\begin{table}[t]
    \centering
    \caption{Codebook update strategy ablation: top-1 accuracy
             (\%) on CIFAR-100 (ViT-Base/16). Single-device
             baseline: $92.30\%$.}
    \label{tab:ablation_update}
    \renewcommand{\arraystretch}{1.15}
    \begin{tabular}{@{}lcc@{}}
        \toprule
        \textbf{Update strategy} & $P{=}2$ & $P{=}3$  \\
        \midrule
        Gradient-based              & 90.47 & 88.44  \\
        EMA (\name\ default)        & 90.51 & 89.00  \\
        EMA + dim.\ reduction ($d_q{=}256$)
                                    & \textbf{90.79} & \textbf{89.30}  \\
        \bottomrule
    \end{tabular}
\end{table}

\section{Discussion}
\label{sec:discussion}

\paragraph{Implications and limitations}
The results in Section~\ref{sec:eval} indicate that removing the receiver's compute dependency is what drives \name's latency gains over \prism and \astra, and that a single multi-resolution codebook lets a deployed model react to changing link conditions without retraining. This property is likely to matter as much as raw compression across heterogeneous edge devices. One caveat applies: \name is evaluated on encoder-only models, where Section~\ref{sec:eval:ablation} shows that exact buffer precomputability already costs $0.72$--$1.68$ accuracy points.

\paragraph{Extension to decoder-only LLMs}
\name's buffers and split-attention (Eq.~\eqref{eq:split-attn}) assume unconditional cross-device attendance, which holds for the bidirectional encoders evaluated here but not for autoregressive decoder-only LLMs such as Qwen2.5~\cite{yang2025qwen25}: causal masking would require \name's distributed attention mechanism to be made partition- and position-aware, and autoregressive decoding would additionally require a scheduling strategy that allocates compute heterogeneously across devices to manage the differing prefill and decode phases of concurrent requests. These are architectural changes rather than a direct application of the present design, so we did not implement or evaluate \name on decoder-only LLMs in this work and leave this extension to future work.

\section{Conclusion}
\label{sec:conclusion}

This paper presented \name, a communication-efficient framework for distributed Transformer inference across edge devices. \name introduces multi-resolution vector quantization to substantially reduce cross-device communication while consistently improving inference accuracy over the state-of-the-art VQ-based method \astra. The framework further provides two operating variants. \name-nb prioritizes accuracy, achieving up to $1.89\%$ higher accuracy than \astra, whereas \name-b achieves up to $1.45\%$ higher accuracy and up to $1.27\times$ inference speedup over \astra by introducing buffer-compatible quantization to precompute codebook-indexed K/V representations, thereby eliminating receiver-side K/V projection and enabling communication--computation overlap.  In addition, a multi-resolution hierarchical codebook clustering enables runtime bitrate adaptation without retraining. Across ViT-Base/16 and ViViT-B/16 on image and video classification tasks, \name achieves approximately $22{,}624\times$ and $24{,}500\times$ reduction in transmitted activation payload, respectively, relative to uncompressed sequence parallelism. Together, these results demonstrate that \name provides a flexible foundation for communication-efficient distributed Transformer inference, while enabling low-latency execution through precomputed K/V lookup and communication--computation overlap.

\section*{Acknowledgments}
This research was supported by the PANDORA project, funded by the European Union’s Horizon Europe Framework Programme under Grant Agreement No 101135775 and NordForsk Nordic University Cooperation on Edge Intelligence (Grant No. 168043).

\bibliographystyle{IEEEtran}
\bibliography{main}

@inproceedings{46201,title	= {Attention is All You Need},author	= {Ashish Vaswani and Noam Shazeer and Niki Parmar and Jakob Uszkoreit and Llion Jones and Aidan N. Gomez and Lukasz Kaiser and Illia Polosukhin},year	= {2017},URL	= {https://arxiv.org/pdf/1706.03762.pdf}}

@article{dosovitskiy2020image,
  title={An image is worth 16x16 words: Transformers for image recognition at scale},
  author={Dosovitskiy, Alexey and Beyer, Lucas and Kolesnikov, Alexander and Weissenborn, Dirk and Zhai, Xiaohua and Unterthiner, Thomas and Dehghani, Mostafa and Minderer, Matthias and Heigold, Georg and Gelly, Sylvain and others},
  journal={arXiv preprint arXiv:2010.11929},
  year={2020}
}

@inproceedings{arnab2021vivit,
  title={Vivit: A video vision transformer},
  author={Arnab, Anurag and Dehghani, Mostafa and Heigold, Georg and Sun, Chen and Lu{\v{c}}i{\'c}, Mario and Schmid, Cordelia},
  booktitle={Proceedings of the IEEE/CVF international conference on computer vision},
  pages={6836--6846},
  year={2021}
}

@inproceedings{touvron2021training,
  title={Training data-efficient image transformers \& distillation through attention},
  author={Touvron, Hugo and Cord, Matthieu and Douze, Matthijs and Massa, Francisco and Sablayrolles, Alexandre and J{\'e}gou, Herv{\'e}},
  booktitle={International conference on machine learning},
  pages={10347--10357},
  year={2021},
  organization={PMLR}
}

@techreport{krizhevsky2009learning,
  title={Learning multiple layers of features from tiny images},
  author={Krizhevsky, Alex},
  year={2009},
  institution={University of Toronto}
}

@INPROCEEDINGS{5206848,
  author={Deng, Jia and Dong, Wei and Socher, Richard and Li, Li-Jia and Kai Li and Li Fei-Fei},
  booktitle={2009 IEEE Conference on Computer Vision and Pattern Recognition}, 
  title={ImageNet: A large-scale hierarchical image database}, 
  year={2009},
  pages={248-255},
  doi={10.1109/CVPR.2009.5206848}}

@inproceedings{devlin2019bert,
  title={Bert: Pre-training of deep bidirectional transformers for language understanding},
  author={Devlin, Jacob and Chang, Ming-Wei and Lee, Kenton and Toutanova, Kristina},
  booktitle={Proceedings of the 2019 conference of the North American chapter of the association for computational linguistics: human language technologies, volume 1 (long and short papers)},
  pages={4171--4186},
  year={2019}
}

@misc{yang2025qwen25,
      title={Qwen2.5 Technical Report}, 
      author={Qwen and : and An Yang and Baosong Yang and others},
      year={2025},
      eprint={2412.15115},
      archivePrefix={arXiv},
      primaryClass={cs.CL},
      url={https://arxiv.org/abs/2412.15115}, 
}

@article{shoeybi2019megatron,
  title={Megatron-lm: Training multi-billion parameter language models using model parallelism},
  author={Shoeybi, Mohammad and Patwary, Mostofa and Puri, Raul and LeGresley, Patrick and Casper, Jared and Catanzaro, Bryan},
  journal={arXiv preprint arXiv:1909.08053},
  year={2019}
}

@article{fang2024usp,
  title={Usp: A unified sequence parallelism approach for long context generative ai},
  author={Fang, Jiarui and Zhao, Shangchun},
  journal={arXiv preprint arXiv:2405.07719},
  year={2024}
}

@article{pope2023efficiently,
  title={Efficiently scaling transformer inference},
  author={Pope, Reiner and Douglas, Sholto and Chowdhery, Aakanksha and Devlin, Jacob and Bradbury, James and Heek, Jonathan and Xiao, Kefan and Agrawal, Shivani and Dean, Jeff},
  journal={Proceedings of machine learning and systems},
  volume={5},
  pages={606--624},
  year={2023}
}

@inproceedings{anthony2024demystifying,
  title={Demystifying the communication characteristics for distributed transformer models},
  author={Anthony, Quentin and Michalowicz, Benjamin and Hatef, Jacob and Xu, Lang and Abduljabbai, Mustafa and Shafi, Aamir and Subramoni, Hari and Panda, Dhabaleswar K},
  booktitle={2024 IEEE Symposium on High-Performance Interconnects (HOTI)},
  pages={57--65},
  year={2024},
  organization={IEEE}
}

@INPROCEEDINGS{10546617,
  author={Wei, Yuanxin and Ye, Shengyuan and Jiang, Jiazhi and Chen, Xu and Huang, Dan and Du, Jiangsu and Lu, Yutong},
  booktitle={2024 Design, Automation \& Test in Europe Conference \& Exhibition (DATE)}, 
  title={Communication-Efficient Model Parallelism for Distributed In-Situ Transformer Inference}, 
  year={2024},
  volume={},
  number={},
  pages={1-6},
  doi={10.23919/DATE58400.2024.10546617}}

@article{huang2019gpipe,
  title={Gpipe: Efficient training of giant neural networks using pipeline parallelism},
  author={Huang, Yanping and Cheng, Youlong and Bapna, Ankur and Firat, Orhan and Chen, Dehao and Chen, Mia and Lee, HyoukJoong and Ngiam, Jiquan and Le, Quoc V and Wu, Yonghui and others},
  journal={Advances in neural information processing systems},
  volume={32},
  year={2019}
}

@inproceedings{li-etal-2023-sequence,
    title = "Sequence Parallelism: Long Sequence Training from System Perspective",
    author = "Li, Shenggui  and
      Xue, Fuzhao  and
      Baranwal, Chaitanya  and
      Li, Yongbin  and
      You, Yang",
    editor = "Rogers, Anna  and
      Boyd-Graber, Jordan  and
      Okazaki, Naoaki",
    booktitle = "Proceedings of the 61st Annual Meeting of the Association for Computational Linguistics (Volume 1: Long Papers)",
    month = jul,
    year = "2023",
    address = "Toronto, Canada",
    publisher = "Association for Computational Linguistics",
    url = "https://aclanthology.org/2023.acl-long.134/",
    doi = "10.18653/v1/2023.acl-long.134",
    pages = "2391--2404"
}

@INPROCEEDINGS{10631032,
  author={Hu, Chenghao and Li, Baochun},
  booktitle={2024 IEEE 44th International Conference on Distributed Computing Systems (ICDCS)}, 
  title={When the Edge Meets Transformers: Distributed Inference with Transformer Models}, 
  year={2024},
  volume={},
  number={},
  pages={82-92},
  doi={10.1109/ICDCS60910.2024.00017}}

@article{li2025tpi,
  title={TPI-LLM: Serving 70B-scale LLMs efficiently on low-resource mobile devices},
  author={Li, Zonghang and Feng, Wenjiao and Guizani, Mohsen and Yu, Hongfang},
  journal={IEEE Transactions on Services Computing},
  year={2025},
  publisher={IEEE}
}

@ARTICLE{10818760,
  author={Zhang, Mingjin and Shen, Xiaoming and Cao, Jiannong and Cui, Zeyang and Jiang, Shan},
  journal={IEEE Internet of Things Journal}, 
  title={EdgeShard: Efficient LLM Inference via Collaborative Edge Computing}, 
  year={2025},
  volume={12},
  number={10},
  pages={13119-13131},
  doi={10.1109/JIOT.2024.3524255}}

@inproceedings{jiang2023hexgen,
author = {Jiang, Youhe and Yan, Ran and Yao, Xiaozhe and Zhou, Yang and Chen, Beidi and Yuan, Binhang},
title = {HEXGEN: generative inference of large language model over heterogeneous environment},
year = {2024},
publisher = {JMLR.org},
booktitle = {Proceedings of the 41st International Conference on Machine Learning},
articleno = {881},
numpages = {16},
location = {Vienna, Austria},
series = {ICML'24}
}

@inproceedings{guo2023sti,
  title={Sti: Turbocharge nlp inference at the edge via elastic pipelining},
  author={Guo, Liwei and Choe, Wonkyo and Lin, Felix Xiaozhu},
  booktitle={Proceedings of the 28th ACM International Conference on Architectural Support for Programming Languages and Operating Systems, Volume 2},
  pages={791--803},
  year={2023}
}

@inproceedings{ye2025jupiter,
  title={Jupiter: Fast and resource-efficient collaborative inference of generative llms on edge devices},
  author={Ye, Shengyuan and Ouyang, Bei and Zeng, Liekang and Qian, Tianyi and Chu, Xiaowen and Tang, Jian and Chen, Xu},
  booktitle={IEEE INFOCOM 2025-IEEE Conference on Computer Communications},
  pages={1--10},
  year={2025},
  organization={IEEE}
}

@article{peccia2024embedded,
  title={Embedded Distributed Inference of Deep Neural Networks: A Systematic Review},
  author={Peccia, Federico Nicol{\'a}s and Bringmann, Oliver},
  journal={arXiv preprint arXiv:2405.03360},
  year={2024}
}

@INPROCEEDINGS{10446903,
  author={Cheng, Yujun and Zhang, Zhewei and Wang, Shengjin},
  booktitle={ICASSP 2024 - 2024 IEEE International Conference on Acoustics, Speech and Signal Processing (ICASSP)}, 
  title={RCIF: Towards Robust Distributed DNN Collaborative Inference Under Highly Lossy Networks}, 
  year={2024},
  volume={},
  number={},
  pages={5350-5354},
  doi={10.1109/ICASSP48485.2024.10446903}}

@inproceedings{
liu2025communication,
title={{ASTRA}: Communication-Efficient Acceleration for Multi-Device Transformer Inference},
author={Xiao Liu and Lijun Zhang and Deepak Ganesan and Hui Guan},
booktitle={Forty-third International Conference on Machine Learning},
year={2026},
url={https://openreview.net/forum?id=QmiKRq3e4N}
}

@ARTICLE{9459473,
  author={Mitra, Aritra and Richards, John A. and Bagchi, Saurabh and Sundaram, Shreyas},
  journal={IEEE Transactions on Signal Processing}, 
  title={Distributed Inference With Sparse and Quantized Communication}, 
  year={2021},
  volume={69},
  number={},
  pages={3906-3921},
  doi={10.1109/TSP.2021.3089824}}

@book{10.5555/128857,
author = {Gersho, Allen and Gray, Robert M.},
title = {Vector quantization and signal compression},
year = {1991},
isbn = {0792391810},
publisher = {Kluwer Academic Publishers},
address = {USA}
}

@article{van2017neural,
  title={Neural discrete representation learning},
  author={Van Den Oord, Aaron and Vinyals, Oriol and others},
  journal={Advances in neural information processing systems},
  volume={30},
  year={2017}
}

@article{yang2023hifi,
  title={Hifi-codec: Group-residual vector quantization for high fidelity audio codec},
  author={Yang, Dongchao and Liu, Songxiang and Huang, Rongjie and Tian, Jinchuan and Weng, Chao and Zou, Yuexian},
  journal={arXiv preprint arXiv:2305.02765},
  year={2023}
}

@article{soomro2012ucf101,
  title={Ucf101: A dataset of 101 human actions classes from videos in the wild},
  author={Soomro, Khurram and Zamir, Amir Roshan and Shah, Mubarak},
  journal={arXiv preprint arXiv:1212.0402},
  year={2012}
}

@book{kaufman2009finding,
  title={Finding groups in data: an introduction to cluster analysis},
  author={Kaufman, Leonard and Rousseeuw, Peter J},
  year={2009},
  publisher={John Wiley \& Sons}
}

@article{qazi2025prism,
  title={PRISM: Distributed Inference for Foundation Models at Edge},
  author={Qazi, Muhammad Azlan and Iosifidis, Alexandros and Zhang, Qi},
  journal={arXiv preprint arXiv:2507.12145},
  year={2025}
}

@article{touvron2023llama,
  title={Llama: Open and efficient foundation language models},
  author={Touvron, Hugo and Lavril, Thibaut and Izacard, Gautier and Martinet, Xavier and Lachaux, Marie-Anne and Lacroix, Timoth{\'e}e and Rozi{\`e}re, Baptiste and Goyal, Naman and Hambro, Eric and Azhar, Faisal and others},
  journal={arXiv preprint arXiv:2302.13971},
  year={2023}
}

@article{touvron2023llama2,
  title={Llama 2: Open foundation and fine-tuned chat models},
  author={Touvron, Hugo and Martin, Louis and Stone, Kevin and Albert, Peter and Almahairi, Amjad and Babaei, Yasmine and Bashlykov, Nikolay and Batra, Soumya and Bhargava, Prajjwal and Bhosale, Shruti and others},
  journal={arXiv preprint arXiv:2307.09288},
  year={2023}
}

@article{guo2025deepseek,
  author = {Guo, Daya and Yang, Dejian and Zhang, Haowei and Song, Junxiao and Wang, Peiyi and Zhu, Qihao and Xu, Runxin and Zhang, Ruoyu and Ma, Shirong and Bi, Xiao and others},
  title = {DeepSeek-R1 incentivizes reasoning in LLMs through reinforcement learning},
  journal = {Nature},
  year = {2025},
  volume = {645},
  number = {8081},
  pages = {633--638},
  month = sep,
  doi = {10.1038/s41586-025-09422-z},
  url = {https://doi.org/10.1038/s41586-025-09422-z},
  issn = {1476-4687}
}

@misc{google2025veo3launch,
  author       = {Google},
  title        = {Veo 3 Launch},
  howpublished = {\url{https://cloud.google.com/blog/products/ai-machine-learning/veo-3-fast-available-for-everyone-on-vertex-ai}},
  year         = {2025},
  note         = {Accessed: September 28, 2025}
}

\end{document}